\documentclass{article}

\PassOptionsToPackage{numbers, compress}{natbib}

\usepackage{wrapfig}

\usepackage[final]{neurips_2025}

\usepackage[final]{changes}
\usepackage[utf8]{inputenc} 
\usepackage[T1]{fontenc}    
\usepackage[hidelinks]{hyperref}
\usepackage{url}            
\usepackage{booktabs}       
\usepackage{amsfonts}       
\usepackage{nicefrac}       
\usepackage{microtype}      
\usepackage{xcolor}         
\usepackage[numbers]{natbib}

\usepackage{comment}
\usepackage{graphicx}
\usepackage{multirow}

\usepackage{xcolor, colortbl, booktabs}
\usepackage{caption}
\definecolor{red1}{HTML}{FFF5F5}  
\definecolor{red2}{HTML}{FCDADA}  
\definecolor{red3}{HTML}{F8BEBE}  
\definecolor{red4}{HTML}{F29494}  
\definecolor{red5}{HTML}{EB7070}  
\definecolor{red6}{HTML}{D64545}  
\usepackage{subcaption}
\usepackage{pifont}
\usepackage{xcolor}
\usepackage[most]{tcolorbox}
\tcbuselibrary{listingsutf8}

\newtcolorbox{promptbox}[1][]{%
  colback=gray!5, 
  colframe=gray!30, 
  fonttitle=\normalfont,
  coltitle=black,
  title=Prompt,
  sharp corners,
  boxrule=0.7pt,
  breakable,
  enhanced,
  boxsep=5pt,
  #1
}

\newcommand{\cmark}{\textcolor{green!60!black}{\ding{51}}} 
\newcommand{\xmark}{\textcolor{red!75!black}{\ding{55}}}   

\title{Worse than Zero-shot? \\ A Fact-Checking Dataset for Evaluating the Robustness of RAG Against Misleading Retrievals}

\author{%
  Linda Zeng\thanks{Equal contribution.} \\
  The Harker School \\
  San Jose, California, USA \\
  \texttt{lindazeng979@gmail.com}
  \And
  Rithwik Gupta\footnotemark[1] \\
  Irvington High School \\
  Fremont, California, USA \\
  \texttt{rithwikca2020@gmail.com}
  \And
  Divij Motwani \\
  Palo Alto High School \\
  Palo Alto, California, USA \\
  \texttt{divijmotwani@gmail.com}
  \And
  Yi Zhang\thanks{Co-advising.} \\
  University of California Santa Cruz \\
  Santa Cruz, California, USA \\
  \texttt{yiz@ucsc.edu}
  \And
  Diji Yang\footnotemark[2] \\
  University of California Santa Cruz \\
  Santa Cruz, California, USA \\
  \texttt{dyang39@ucsc.edu}
}

\begin{document}

\maketitle

\begin{abstract}
Retrieval-augmented generation (RAG) has shown impressive capabilities in mitigating hallucinations in large language models (LLMs). However, LLMs struggle to maintain consistent reasoning when exposed to misleading or conflicting evidence, especially in real-world domains such as politics, where information is polarized or selectively framed. 
Mainstream RAG benchmarks evaluate models under clean retrieval settings, where systems generate answers from gold-standard documents, or under synthetically perturbed settings, where documents are artificially injected with noise. These assumptions fail to reflect real-world conditions, often leading to an overestimation of RAG system performance. To address this gap, we introduce \textsc{RAGuard}, the first benchmark to evaluate the robustness of RAG systems against \textit{misleading} retrievals. Unlike prior benchmarks that rely on synthetic noise, our fact-checking dataset captures naturally occurring misinformation by constructing its retrieval corpus from Reddit discussions. It categorizes retrieved evidence into three types: \textit{supporting}, \textit{misleading}, and \textit{unrelated}, providing a realistic and challenging testbed for assessing how well RAG systems navigate different types of evidence. Our experiments reveal that, when exposed to potentially misleading retrievals, all tested LLM-powered RAG systems perform worse than their zero-shot baselines (i.e., no retrieval at all), while human annotators consistently perform better, highlighting LLMs' susceptibility to noisy environments. To our knowledge, \textsc{RAGuard} is the first benchmark to systematically assess the robustness of the RAG against misleading evidence.
We expect this benchmark to drive future research toward improving RAG systems beyond idealized datasets, making them more reliable for real-world applications.\setcounter{footnote}{0}\footnote{The dataset is available at \href{https://huggingface.co/datasets/UCSC-IRKM/RAGuard}{https://huggingface.co/datasets/UCSC-IRKM/RAGuard}.}
\end{abstract}

\section{Introduction}

\begin{figure}
    \centering

    \includegraphics[width=\linewidth]{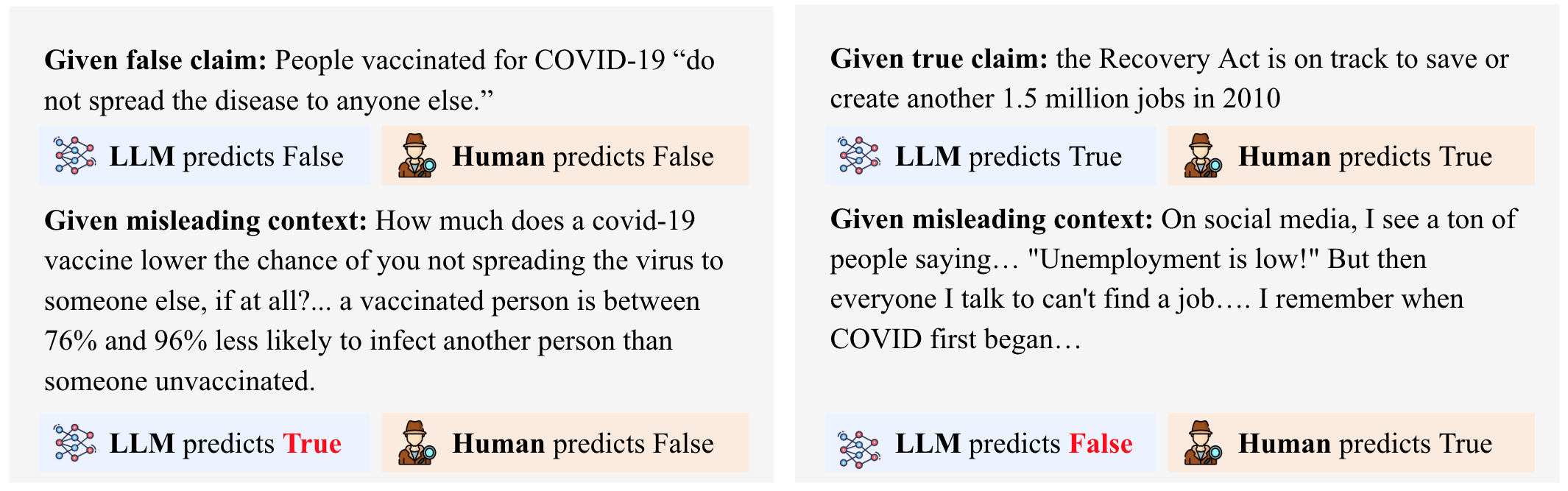}
    \caption{Examples of LLM and human performance on a false claim (left) and a true claim (right) from \textsc{RAGuard}. While the LLM initially classified both claims correctly, it later reversed its decisions due to misleading retrieved context. In contrast, human judgments remained consistent.}
   
    \label{fig:datasetexample}
\end{figure}

Retrieval-augmented generation (RAG) systems show strong potential for mitigating large language model (LLM) hallucinations and enhancing trustworthiness. By combining the generative capabilities of LLMs with the retrieval power of external corpora, RAG aims to ground responses in relevant information, thus improving factual consistency and output credibility \cite{RAGNLP,RAGPretrain,gao2023retrieval}. While existing work has made significant progress in improving retrieval relevance and maximizing the amount of information in the retrieved context \cite{karpukhin-etal-2020-dense, yang2024rag, borgeaud2022improving}, \added{comparatively less attention has been paid to scenarios where LLMs must reason over inevitably misleading, conflicting, or only partially relevant retrieved content, as illustrated in Figure~\ref{fig:datasetexample}.} Addressing this robustness gap is increasingly important as RAG systems are deployed in high-stakes applications such as fact-checking \cite{thorne-vlachos-2018-automated} and legal or medical domains \cite{guha2024legalbench, xiong2024benchmarkingretrievalaugmentedgenerationmedicine}.


Prior work has mitigated LLMs' susceptibility to noisy retrievals by prompting models to evaluate each retrieved document's relevance \cite{InstructRAG}, isolating LLMs' responses to individual passages before aggregating them \cite{RobustRAG}, or prompting agents to select external knowledge based on a debate process \cite{LearningToBreak}. However, these approaches largely focus on filtering or restructuring retrieval rather than tackling the core challenge of LLM reasoning over misleading information. \added{Many approaches aim to reconcile temporal or factual inconsistencies between retrieved content and an LLM’s prior knowledge rather than addressing cases where the retrieved information itself is misleading or contradictory} \cite{AstuteRAG, ContextMemoryConflict}. Furthermore, current datasets overly rely on curating reliable documents, limiting robustness testing against misinformation \cite{TriviaQA, HotPotQA, SQUAD, NaturalQuestions}. While some introduce counterfactuals or retrieval noise \cite{qacc, PowerOfNoise}, they rely on artificial perturbations or costly human annotation. This highlights the need for an evaluation framework that challenges RAG systems with naturally occurring contradictions, \added{as well as a taxonomy of evidence types that clarifies their distinct impacts on model behavior. }


Fact-checking plays a crucial role in combating misinformation, yet most existing datasets \added{in this domain} assume the availability of gold-standard evidence aligning with the verdict \cite{FEVER,feverous,mocheg,snopes,pubhealth,multifc}. 
This assumption breaks down in political domains, where controversial claims lead to both supporting and opposing narratives from diverse sources \cite{nielsen2022mumin, FEVER, PolitifactOslo, FakeNewsNet}. To build fact-checking systems capable of handling real-world misinformation, it is essential to expose models to the conflicting and misleading evidence with which humans work in the real world. 


To bridge this gap, we introduce \textsc{RAGuard}, a benchmark dataset based on political claims and their verifications from PolitiFact that incorporates real-world misinformation. Given the prevalence of polarizing and deceptive information in political discourse, we develop an automated pipeline that retrieves relevant yet potentially misleading user-generated content and labels them through a novel LLM-guided approach simulating a fact-checking exam. 
We then evaluate widely used LLM-based RAG systems,
confirming that current methods lack robustness in real-world scenarios. Performance drops significantly when RAG systems are exposed to documents from the \textsc{RAGuard} knowledge base, revealing a substantial gap from human reasoning in identifying and handling misleading evidence.

In summary, our work advocates for a shift from idealized RAG settings to ones that reflect the misleading nature of real-world retrieval. We introduce a \added{taxonomy,} benchmark, and evaluation framework for assessing LLM robustness under such conditions. Our contributions are as follows:

\begin{itemize}
\item \textbf{Task:} We define a new robustness-focused fact verification task that challenges models to reason through misleading retrieved content. We also unify inconsistent terminology in prior work by establishing a structured framework for labeling document types (e.g., supporting, misleading, unrelated).

\item \textbf{Benchmark:} We release \textsc{RAGuard}, a real-world political-domain RAG benchmark built from PolitiFact claims and Reddit retrievals. Documents are labeled based on their effect on LLM predictions using a scalable, LLM-guided annotation method targeting misleading effect on models rather than humans.

\item \textbf{Evaluation:} We evaluate strong closed- and open-source models across multiple retrieval settings, revealing that even top-performing LLMs struggle under misleading context with all models performing worse than their zero-context baselines.
\end{itemize}

\section{Dataset}

We introduce \textsc{RAGuard}, a benchmark for evaluating the robustness of RAG systems in political fact-checking.
In the following sections, we standardize terminology used in prior work (Section~\ref{sec:task}), compare \textsc{RAGuard} to existing datasets (Section~\ref{sec:comparison}), describe its structure and key statistics (Section~\ref{sec:statistics}), and outline the fact-checking tasks it supports (Section~\ref{sec:tasks}).


\subsection{Task and Terminology}\label{sec:task}

 The core task in \textsc{RAGuard} is retrieval-augmented fact-checking: determining whether a claim is true or false based on retrieved evidence that may be \textit{supporting}, \textit{misleading}, or \textit{unrelated}. Unlike prior datasets that include only documents explicitly supporting the correct answer, \textsc{RAGuard} adopts a broader definition of \textit{supporting}, allowing documents that provide contextual cues even if they do not state the answer outright.

To better reflect real-world retrieval conditions, we introduce noise through \textit{unrelated} documents, which are topically related but uninformative, and crucially, \textit{misleading} documents, which subtly distort facts through framing, omission, or biased presentation. Unlike adversarially fabricated content or clearly one-sided, unambiguous evidence, misleading noise arises naturally and rarely contains explicit falsehoods; instead, it subtly presents facts or context in ways that are misleading to models but often recognizable to humans (see Figure~\ref{fig:datasetexample}). Importantly, the quality of being misleading in \textsc{RAGuard} is defined relative to language models rather than as an objective property discernible by all humans. The dataset is designed to expose specific vulnerabilities in model reasoning, cases where LLMs fail to separate factual content from bias or rhetorical tone. Thus, the task reflects the more realistic challenge of distilling truth from partial or polarized viewpoints, as human fact-checkers must do, rather than merely answer matching with given evidence.


Finally, we unify terminology from prior work to clarify distinctions between evidence types. While some documents may be non-conflicting but distracting (e.g., unrelated or randomly-selected), others are conflicting and more challenging, such as misleading, fabricated, or unambiguous evidence. We illustrate this hierarchical structure of definitions in Figure~\ref{fig:definition} and explicitly define all document types in our taxonomy (see Appendix~\ref{sec:terminology}), which helps situate \textsc{RAGuard} within the broader space of retrieval-augmented fact-checking datasets.

\subsection{Comparison with Existing Datasets}\label{sec:comparison}

\paragraph{Fact-Checking Datasets.}

\begin{figure*}[t]
\centering
\begin{minipage}[t]{0.58\textwidth}
    \centering
    \setlength{\tabcolsep}{4pt}
    \renewcommand{\arraystretch}{0.9}
    \vspace{0.5em}
    \resizebox{\textwidth}{!}{%
    \begin{tabular}{l| c c|  c c c| c c}
    \toprule
    
    \textbf{Dataset} &
    \multicolumn{2}{c|}{\textbf{Focus}} &
    \multicolumn{3}{c|}{\textbf{Evidence}} &   \multicolumn{2}{c}{\textbf{Claims}}  \\
& FC & ROB & Retrieval & Conflicting & Real-world & Domain & \# Claims \\
        \midrule
        FEVER \cite{FEVER}       & \cmark & \xmark & \cmark& \xmark & \xmark & General   & 185K \\
     FEVEROUS \cite{feverous} & \cmark & \xmark& \cmark & \xmark & \xmark & General   & 87K \\
        Liar \cite{Liar}         & \cmark & \xmark& \xmark & \xmark & \cmark & Political & 12.8K \\
        Mocheg \cite{mocheg}     & \cmark & \xmark& \cmark & \xmark & \cmark & Political & 15.6K \\
        Snopes \cite{snopes}     & \cmark &\xmark& \cmark &  \xmark & \cmark & Political & 6.4K \\
         PubHealth \cite{pubhealth} & \cmark &\xmark& \cmark &  \xmark & \cmark & Health & 11.8K \\
         MultiFC \cite{multifc}   & \cmark & \xmark& \cmark & \xmark & \cmark & Political & 43.8K \\
         AVeriTeC \cite{schlichtkrull2023averitecdatasetrealworldclaim} & \cmark &\xmark& \cmark & \cmark & \cmark & Political & 4.6K \\
        \midrule
        Power of Noise \cite{PowerOfNoise} & \xmark & \cmark& \cmark &\xmark & \cmark & General   & 10K \\
         RAAT \cite{raat}                   & \xmark & \cmark& \cmark &\cmark & \xmark & General   & 7.8K \\
         NoiserBench \cite{NoiserBench}     & \xmark & \cmark& \cmark &\cmark & \xmark & General   & 4K \\
         QACC \cite{qacc}                   & \xmark &\cmark & \cmark& \cmark & \cmark & General   & 1.5K \\
        \midrule
        \textsc{RAGuard} (ours)            & \cmark &\cmark& \cmark & \cmark & \cmark & Political & 2.6K \\
        \bottomrule
    \end{tabular}
    }
    \captionof{table}{Comparison of \textsc{RAGuard} with fact-checking and noisy RAG datasets. ``FC'' indicates suitability for fact-checking, and ``ROB'' for LLM robustness evaluation. Columns reflect evaluation focus, evidence types, and dataset characteristics.}   
    \label{tab:datasets-table}
\end{minipage}
\hfill
\begin{minipage}[t]{0.4\textwidth}
    \centering
          \vspace{1.2em}
          \includegraphics[width=\textwidth]{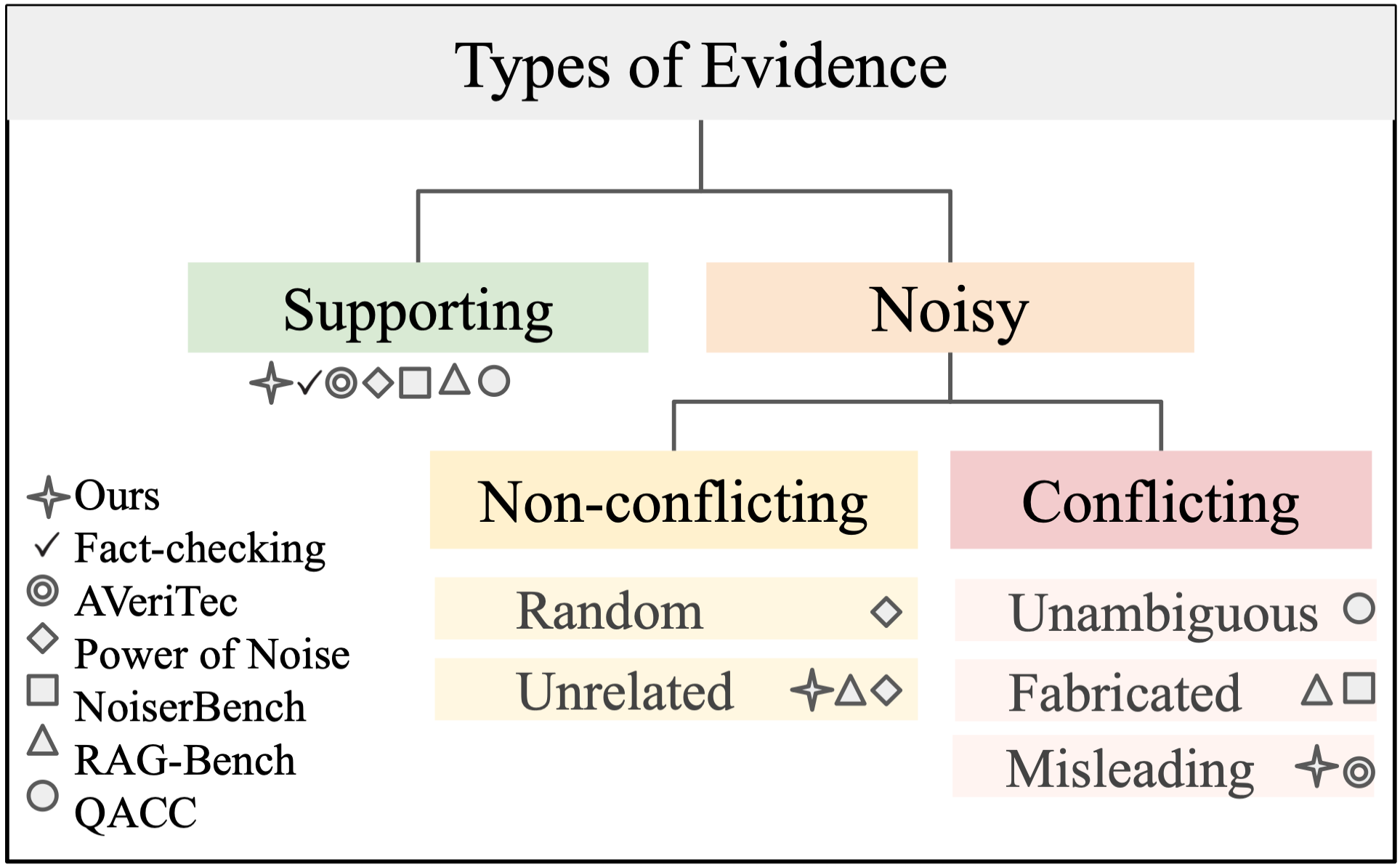}
    \captionof{figure}{Taxonomy of document types used in our benchmark (supporting, misleading, unrelated), along with types of evidence included in related datasets.}

    \label{fig:definition}
\end{minipage}
\end{figure*}

Table~\ref{tab:datasets-table} summarizes key properties of existing fact-checking and RAG benchmarks, including whether retrieval is used, conflicting evidence is included, and documents are drawn from real-world sources. Most fact-checking datasets that incorporate evidence retrieval are limited to supporting documents and do not account for conflicting or misleading information. For example, while FEVEROUS \cite{feverous} categorizes some evidence as \textit{refuted}, this label only applies to documents that help the model correctly classify a claim as false, not those that contradict the fact-checking verdict. 
Additionally, both FEVER \cite{FEVER} and FEVEROUS rely on curated, rewritten Wikipedia passages, rather than naturally occurring claims or user-generated content \cite{multifc}.

While Liar \cite{Liar} and Mocheg \cite{mocheg} also source claims from PolitiFact, Liar does not support evidence retrieval, and Mocheg includes only gold-standard documents cited by PolitiFact fact-checkers. Similarly, other datasets \cite{snopes, pubhealth, multifc} primarily use journalist-written explanations from fact-checking websites, which are explicitly curated to justify the verdict. 
In contrast, \textsc{RAGuard} incorporates conflicting evidence from naturally occurring Reddit discussions, reflecting more realistic challenges.

AVeriTeC \cite{schlichtkrull2023averitecdatasetrealworldclaim} also explores real-world misleading information, focusing on retrieval-augmented fact verification with naturally occurring conflicting evidence. Like \textsc{RAGuard}, their dataset includes claims and evidence drawn from real-world sources, some of which subtly distort the truth through biased framing or misinformation. However, a key difference lies in the task framing: their benchmark treats evidence as ``conflicting'' when it contains both supporting and refuting signals and asks the model to abstain from a definitive verdict in such cases. In contrast, \textsc{RAGuard} explicitly expects the model to reason \textit{through} misleading evidence and arrive at a correct verdict. While AVeriTeC focuses on detecting ambiguity in the evidence, \textsc{RAGuard} challenges models to exhibit robustness under noisy, real-world conditions. As such, our task is not only stricter but also aligned with practical fact-checking requirements.

\paragraph{Datasets with Noisy Contexts.}

Several prior datasets introduce noisy contexts for evaluating retrieval-augmented generation, primarily in open-domain QA settings \cite{qacc,PowerOfNoise,NoiserBench,raat}. However, these datasets vary significantly in how they define noise and the types of disruptions they model. Power of Noise \cite{PowerOfNoise} introduces noise through unrelated documents, in the form of off-target retrievals or random documents entirely irrelevant to the query. Importantly, the dataset does not introduce any content that actively contradicts or distorts the correct answer (i.e., its noise is exclusively non-conflicting.) RAAT \cite{raat} introduces counterfactual noise by editing documents to contain incorrect answers. While this creates explicit conflicting evidence, it does so synthetically, often resulting in unrealistic or adversarial examples that lack subtlety (e.g., ``Titanic earned a worldwide total of 2.187 billion'' is directly replaced with a different number, which would likely mislead not only an LLM but also a human). Similarly, the noise in NoiserBench \cite{NoiserBench} is fabricated, making it poorly representative of how misleading evidence actually appears in public discourse. QACC \cite{qacc} uses human annotators to label naturally retrieved documents as conflicting or not, avoiding artificial edits. However, the definition of conflict in QACC is binary (i.e., whether the document directly supports or contradicts the gold answer) leaving little room for subtler forms of misleading reasoning (see Figure~\ref{fig:datasetexample}).

\textsc{RAGuard} complements these datasets by focusing specifically on real-world political claims, including user-generated texts containing naturalistic misinformation that challenge LLMs with plausible distortions rather than  clear factual opposition. Unlike prior work, which avoids conflicting content or reduces it to fabricated or unambiguous contradictions, \textsc{RAGuard} expects models to reason through \textit{misleading} evidence and infer the correct label, reflecting real-world fact-checking scenarios, where misinformation is embedded in discourse rather than stated outright.

\begin{figure*}[t]
    \centering
    \begin{subfigure}{0.3\linewidth}
        \centering
        \scriptsize
        \renewcommand{\arraystretch}{0.5}
        \begin{tabular}{l@{\hskip 1.5pt}r}
            \toprule
            \textbf{Statistic} & \textbf{Value} \\
            \midrule
            Total Claims & 2,648 \\
            \quad True & 1,333 (50.3\%) \\
            \quad False & 1,315 (49.7\%) \\
            Avg. Claim Length & 17.6 words \\
            \midrule
            Total Documents & 16,331 \\
            \quad Supporting & 2,685 (16.4\%) \\
            \quad Misleading & 1,812 (11.1\%) \\
            \quad Unrelated & 11,834 (72.5\%) \\
            Avg. Doc Length & 161 words \\
            Avg. Docs/Claim & 6.2 \\
            \midrule
            Claims w/ Supporting Docs & 955 (36.1\%) \\
            Claims w/ Misleading Docs & 788 (29.8\%) \\
            \bottomrule
        \end{tabular}
        \caption{Key statistics on claims and documents, including class balance and average lengths.}
        \label{tab:stats}
    \end{subfigure}
    \hfill
    \begin{subfigure}{0.35\linewidth}
        \centering
        \includegraphics[width=\linewidth]{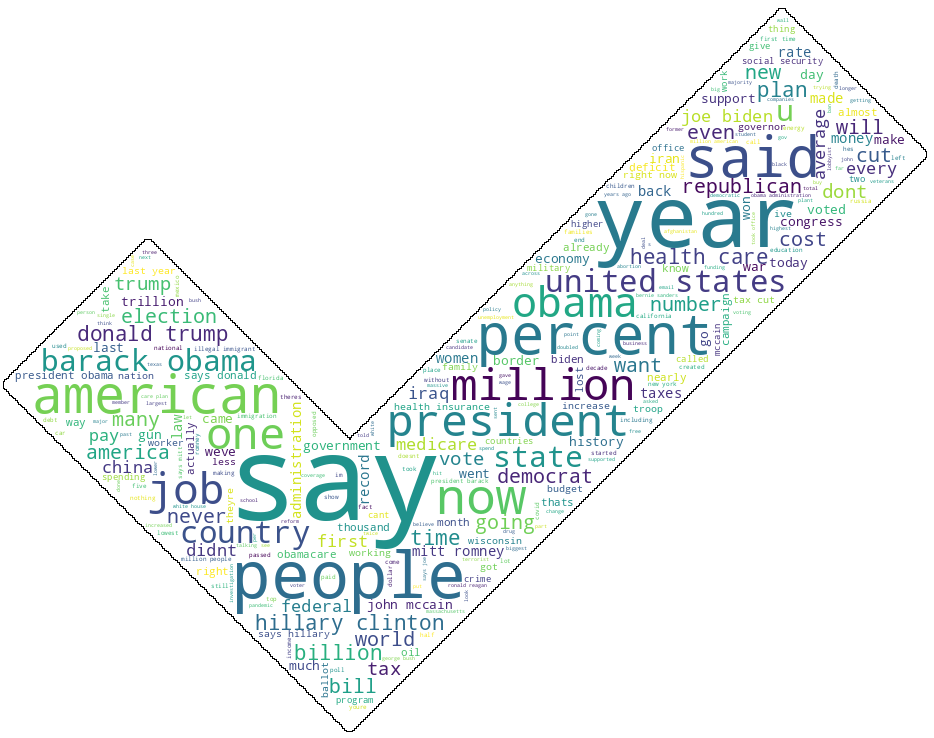}
        \caption{Frequent words in claims, shaped as a checkmark to reflect verification focus.}
        \label{fig:claim_chart}
    \end{subfigure}
    \hfill
    \begin{subfigure}{0.29\linewidth}
        \centering
        \includegraphics[width=\linewidth]{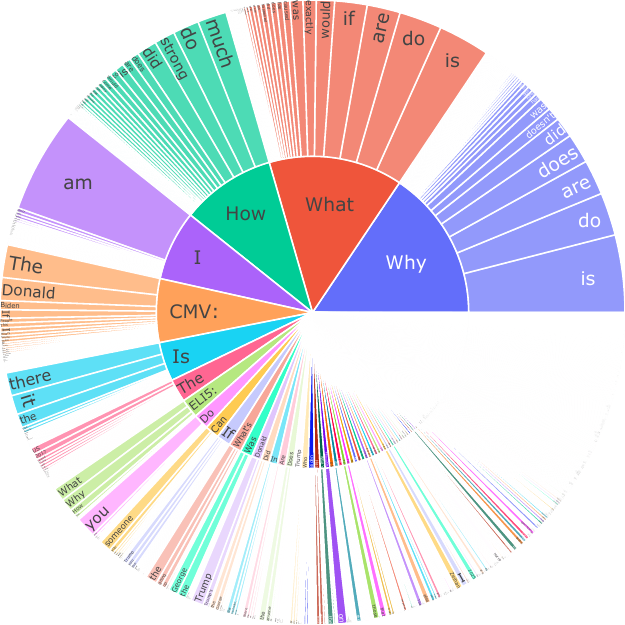}
        \caption{Common opening words in documents, with many thin segments indicating high diversity.}
        \label{fig:document_chart}
    \end{subfigure}
    
    \caption{Overview of \textsc{RAGuard}, including dataset statistics and word frequencies.}
    \label{fig:combined-distribution}
\end{figure*}

\subsection{Dataset Structure}\label{sec:statistics}

\textsc{RAGuard} consists of 2,648 political claims made by U.S. presidential candidates (2000–2024), each labeled as either \textit{true} or \textit{false}, and a knowledge base comprising 16,331 documents. Figure~\ref{tab:stats} presents the key statistics of the dataset. Each claim is linked to a set of associated documents, categorized as \textit{supporting}, \textit{misleading}, or \textit{unrelated}, with an average of 6.2 documents per claim. Notably, the dataset contains more supporting documents than misleading ones, reflecting that political discussions online are more often aligned with factual information, while the large number of unrelated documents suggests that many discussions online are neutral, neither misleading nor supporting the validity of a claim. Appendix~\ref{sec:stats} provides additional statistics related to the year and speaker of each claim. 

\textsc{RAGuard} includes a diverse and realistic collection of political claims and documents. Figure~\ref{fig:claim_chart} visualizes the most frequent words in claims, revealing a focus on reported assertions (e.g., “say”) and quantitative language (e.g., “percent,” “million”) that require quantitative reasoning. The frequent occurrence of temporal terms like “year” further indicates that many claims are time-sensitive, requiring temporal awareness that may challenge both human fact-checkers and LLMs.

Figure~\ref{fig:document_chart} illustrates the lexical diversity of the retrieval corpus. The inner ring denotes the first word in each document’s opening sequence, while the outer ring shows the subsequent word. The abundance of narrow, evenly distributed segments demonstrates that no single phrase or construction dominates the corpus. This diversity helps prevent models from exploiting superficial lexical cues, ensuring that success depends on genuine reasoning rather than memorized linguistic patterns. Notably, many retrieved documents begin with questions, mirroring the exploratory and uncertain tone of real-world online discussions where factuality is often debated rather than asserted.


\subsection{Supported Tasks}\label{sec:tasks}

To benchmark the performance of current RAG systems in real-world fact-checking scenarios, we define a series of tasks using \textsc{RAGuard}. 

\paragraph{Zero-Context Prediction.} This task assesses the model’s ability to fact-check claims using only its internal knowledge, with no retrieved documents. It serves as a zero-shot baseline for evaluating the impact of retrieval.


\paragraph{Standard RAG.}

This task requires RAG systems to retrieved documents from the entire dataset corpus in real time. Retrieved context may include ground-truth documents related to the claim (unrelated, supporting, or misleading information) or unrelated documents, simulating noisy retrieval in real-world settings. 


\paragraph{Oracle Retrieval.}
This task isolates the effect of retrieval content quality by bypassing real-time search and directly supplying documents known to be associated with a given claim. We evaluate two conditions: In the first, which we denote as Oracle Retrieval (All), the model receives a document labeled as supporting, misleading, or unrelated, testing its ability to reason over mixed or ambiguous evidence. In the second, which we refer to as Oracle Retrieval (Misleading), the model is exposed only to documents that conflict with the claim’s ground-truth label, providing a targeted evaluation of susceptibility to deceptive or adversarial content.

\section{Dataset Construction}
\label{gen_inst}

\begin{figure*}
    \centering
  
    \includegraphics[width=\textwidth]{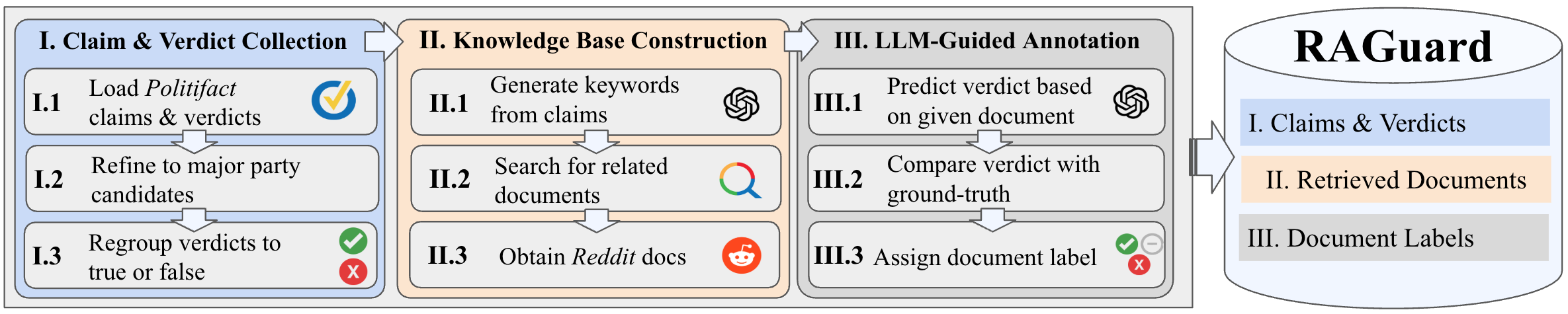} 
      \caption{\textsc{RAGuard} dataset construction, consisting of three stages to obtain claims and verdicts, associated documents, and labels for the each document's relationship to the claim and verdict.}
    \label{fig:construction}
\end{figure*}

\textsc{RAGuard} is constructed in three stages, as shown in Figure~\ref{fig:construction}: (1) collecting political claims and verdicts, (2) retrieving documents from Reddit, and (3) labeling documents as \textit{supporting}, \textit{misleading}, or \textit{unrelated} using an LLM-guided protocol. Our key novelty lies in stage (3), where we define being misleading relative to model behavior rather than human judgment, allowing \textsc{RAGuard} to scale beyond human annotation and target LLM-specific weaknesses.


\subsection{Claim and Verdict Collection}
\label{sec:claim}

We scrape political claims and verdicts from PolitiFact,\footnote{\url{https://www.politifact.com}} a professional fact-checking organization. To ensure claims are both prominent and controversial enough to spur misinformation, we restrict our dataset to US presidential candidates from 2000 to 2024. For clarity in downstream document retrieval and annotation, we binarize PolitiFact's six-point truth scale (\textit{true, mostly true, half true, mostly false, false, pants on fire}) into \textit{true} and \textit{false} verdicts, omitting intermediate categories as it is challenging for a document to specifically mislead a \textit{half true} verdict. 

\subsection{Knowledge Base Construction}
\label{sec:knowledge}

To simulate real-world retrieval noise, we construct a retrieval corpus from Reddit,\footnote{\url{https://www.reddit.com}} a platform rich with user-generated political discussion and misinformation. Because Reddit users inherently publish their opinions to the public, posts are often written to be convincing, regardless of how true the information is. For each claim, we use GPT-4 to generate keyword variants, increasing the likelihood of retrieving diverse types of information related to the topic. We then issue a Google Search restricted to Reddit and collect the top ten retrieved posts per claim. This keyword-based search ensures contextual relevance, while Reddit’s user-generated content introduces diverse perspectives, including both speculative theories and well-supported arguments, mirroring the complexities of real-world fact-checking challenges. 

\subsection{LLM-Guided Annotation}
\label{sec:llm-annotation}

We annotate each retrieved document based on its functional effect on the model’s prediction. Rather than relying on manual judgments or synthetic counterfactuals, we simulate the document’s use in a RAG pipeline by prompting GPT-4 to classify the claim using only the document as context. We then compare its prediction to the ground-truth verdict. If the document helps the model arrive at the correct label, it is labeled \textit{supporting}. If the document causes an incorrect prediction, it is labeled \textit{misleading}. If the model considers the document unrelated to its prediction, it is labeled \textit{unrelated}. Appendix~\ref{sec:prompts1} includes prompts for keyword extraction and LLM annotation.

This LLM-as-annotator strategy enables scalable, behavior-based labeling. Because our dataset aims to reveal weaknesses in how LLMs interpret biased or conflicting information, we define misleadingness relative to model behavior (i.e., by a document’s ability to confuse the model rather than by human interpretation). Human annotators can typically resolve these cases with ease (Section~\ref{sec:human-study}). Moreover, these labels are used solely as an intermediate analytical tool to characterize how retrieved evidence influences model behavior. While models may encounter incorrect or misleading information in the retrieved context, the task evaluates their ability to verify claims against gold verdicts from PolitiFact, rather than to detect misinformation within the documents themselves.

\added{To confirm this process does not overfit to GPT-4 idiosyncrasies, we re-annotated a subset of the dataset with Claude 3.5 Sonnet and Gemini 1.5 Flash and measured inter-annotator agreement using Cohen’s $\kappa$.
We observe substantial agreement with Claude 3.5 Sonnet ($\kappa$ = 0.789) and moderate agreement with Gemini 1.5 Flash ($\kappa$ = 0.650) on the supporting and misleading labels. These results indicate that the labeling decisions are reasonably stable across models rather than solely an artifact of GPT-4’s annotation behavior. As shown in Section~\ref{sec:model-results}, multiple LLMs exhibit similar vulnerabilities on the same misleading documents, reinforcing the generality of our findings.}

\section{Baselines}\label{sec:model-results}

\subsection{Experimental Setup}

\paragraph{Evaluation.} We frame fact-checking as a binary classification task where the model must generate a response that aligns with one of the predefined options. Accuracy, calculated using the ground-truth verdict, is used to evaluate performance. If a model generates an out-of-scope response that does not match any of the given options, it is treated as an incorrect prediction. 

\paragraph{Implementation Details.} 
We evaluate \added{eight LLMs: three open-source models at different scales (OLMo-1B \cite{groeneveld2024olmo},} Llama 3 8B Instruct \cite{dubey2024llama}, and Mistral 7B Instruct \cite{jiang2023mistral7b}), three commercial APIs (Gemini 1.5 Flash \cite{geminiteam2024gemini15unlockingmultimodal}, GPT-4o \cite{openai2024gpt4technicalreport}, and Claude 3.5 Sonnet \cite{claude2024}), and two closed-source reasoning-oriented models (DeepSeek R1 \cite{deepseekai2025deepseekr1incentivizingreasoningcapability} and o4-mini \cite{openai2025o4mini}). For the Standard RAG setting, we perform retrieval using OpenAI’s \texttt{text-embedding-ada-002} model, retrieving the top one (RAG-1) and five (RAG-5) documents based on semantic similarity to the claim. In the Oracle Retrieval setting, we directly supply the pre-labeled associated documents without performing retrieval. All prompts indicate that provided context may be unrelated or factually incorrect (see Appendix~\ref{sec:prompts2}).

\added{In addition to standard baselines, we evaluate a robustness-oriented RAG method, Corrective RAG (CRAG) \cite{yan2024corrective}. We reproduce CRAG using the authors’ released code, prompts, and pretrained Critic. 
All experiments are conducted under our Oracle Retrieval setting, where CRAG receives pre-labeled documents from \textsc{RAGuard}. We compare their Llama 2-based system to zero-shot Llama 2. }

\subsection{Results}

Table~\ref{tab:main} displays baseline results on \textsc{RAGuard} for three tasks using three open-source, three closed-source, and two closed-source reasoning LLMs.

\begin{table*}[t]
\centering
\resizebox{\textwidth}{!}{%
\begin{tabular}{lcccccccc}
\toprule
& \multicolumn{3}{c}{Open Source} & \multicolumn{3}{c}{Closed Source} & \multicolumn{2}{c}{Reasoning} \\
\cmidrule(lr){2-4} \cmidrule(lr){5-7} \cmidrule(lr){8-9}
& OLMo-1B & Llama 3 & Mistral & Gemini 1.5 & GPT-4o & Claude 3.5 & DeepSeek & o4-mini \\
\midrule
Zero-Context Prediction & 
56.87 & 62.50 & 63.97 & 
61.06 & 67.33 & 74.51 & 
69.98 & 63.67 \\

RAG-1 & 
\cellcolor{red1}52.68 & 
\cellcolor{red2}59.40 & 
\cellcolor{red3}59.14 & 
\cellcolor{red3}56.68 & 
\cellcolor{red2}64.80 & 
\cellcolor{red2}70.09 & 
\cellcolor{red2}66.88 & 
\cellcolor{red1}62.76 \\

RAG-5 & 
\cellcolor{red2}49.74 & 
\cellcolor{red1}61.37 & 
\cellcolor{red3}58.91 & 
\cellcolor{red2}57.59 & 
\cellcolor{red1}65.90 & 
\cellcolor{red3}68.58 & 
\cellcolor{red4}57.81 & 
\cellcolor{red1}63.14 \\

Oracle Retrieval (All) & 
\cellcolor{red1}53.89 & 
\cellcolor{red1}61.09 & 
\cellcolor{red4}51.55 & 
\cellcolor{red4}52.38 & 
\cellcolor{red4}53.22 & 
\cellcolor{red5}52.56 & 
\cellcolor{red5}50.06 & 
\cellcolor{red4}51.88 \\

Oracle Retrieval (Misleading) & 
\cellcolor{red4}44.04 & 
\cellcolor{red5}36.81 & 
\cellcolor{red6}26.88 & 
\cellcolor{red6}30.57 & 
\cellcolor{red5}45.97 & 
\cellcolor{red6}35.98 & 
\cellcolor{red6}38.25 & 
\cellcolor{red6}33.39 \\
\bottomrule
\end{tabular}
}
\caption{Accuracy (\%) of various LLM backbones in RAG setup across three tasks and five evaluation settings. Cell color intensity corresponds to the model's percent accuracy drop relative to its zero-context baseline, with darker red indicating larger relative performance drop. Appendix~\ref{sec:analysis} contains a version of Table~\ref{tab:main} with exact relative percent decreases shown.}
\label{tab:main}
\end{table*}

\paragraph{Zero-Context Prediction.}
All systems achieve the highest accuracy on the zero-context prediction (i.e., zero-shot baseline), which is counterintuitive, considering this setting does not benefit from retrieval. Furthermore, we find that reasoning models do not achieve higher zero-context prediction scores since the zero-context task relies primarily on prior knowledge rather than reasoning capability.

\paragraph{Standard RAG.}
Adding retrieved context consistently reduces performance across all models. While o4-mini is the most robust, the performance of models like Mistral and Gemini 1.5 drops sharply. More retrieved documents (RAG-5) often worsen performance compared to RAG-1, especially for stronger models like Claude and DeepSeek, suggesting that retrieval introduces primarily distracting and misleading information, and when the quality of retrieved context is not optimal, including more documents can confuse rather than help, particularly if the model already performs well on the task. These findings challenge the assumption that retrieval improves accuracy and align with concerns about retrieval quality in real-world tasks \cite{relevance, yin2023alcunalargelanguagemodels, PowerOfNoise, xie2024adaptivechameleonstubbornsloth}.

\paragraph{Oracle Retrieval.}
In the Oracle Retrieval (All) condition, where models receive all documents explicitly associated with each claim, performance drops even further compared with both Standard RAG and the zero-shot baseline. The effect is most severe in the Oracle Retrieval (Misleading) condition, which yields an average accuracy decrease of 46.5\%. Every model falls below 50\% accuracy despite the binary nature of the task, confirming that the misleading evidence in \textsc{RAGuard} explicitly disrupts model reasoning. These results demonstrate that current RAG systems are unable to distinguish factual content from subtle misinformation.


\paragraph{Existing Robustness Method.} \added{Despite being designed to mitigate retrieval errors, the CRAG method performs substantially worse than the zero-shot baseline: Llama 2 achieves 50.57\% without retrieval but drops to 37.24\% with CRAG in the Oracle Retrieval Setting. This method uses a lightweight evaluator to score document relevance and, when confidence is low, triggers web searches outside the retrieval corpus, primarily drawing from sources such as Wikipedia. In practice, this mechanism activates in 70.1\% of \textsc{RAGuard} cases, but the added web content often introduces additional noise, as general-purpose sources such as Wikipedia provide limited support for verifying complex political claims. When the evaluator’s confidence is low, CRAG combines web search results with documents from the corpus, effectively doubling exposure to noisy content. Therefore, while CRAG can detect the challenge of the problem, it remains unable to resist the misleading information. More broadly, this underscores that existing robustness methods, which have been shown to handle certain types of noise, remain vulnerable to the qualitatively different challenge of real-world misleading retrieval.}

\subsection{Analysis}
\paragraph{Model Comparison.}
 Across models, GPT-4o is the most robust overall: its accuracy falls by only 31.7\% in the Oracle Retrieval (Misleading) condition, compared to Claude 3.5 (51.8\%), Gemini 1.5 (49.9\%), and Mistral (58.0\%). This pattern seems counterintuitive given that parts of the dataset were shaped by GPT-4’s own failure modes. Nevertheless, its robustness in this setting suggests that the annotation process does not overly capture GPT-4-specific failure patterns and that its high performance may instead be bolstered by its inherent reasoning strength, consistent with recent studies showing GPT-4’s superior fact-checking capabilities~\cite{tang2024minicheck}, as further discussed in Appendix~\ref{sec:gpt4role}.  In constrast, Claude 3.5 achieves the highest zero-context accuracy but suffers the steepest decline, suggesting that strong internal knowledge does not guarantee resistance to misleading information. 
 
 Reasoning-focused systems such as o4-mini and DeepSeek show large performance drops of 45.3\% and 47.6\%, respectively, and \added{smaller models like OLMo-1B follow the same pattern, with all RAG variants underperforming their zero-shot baselines.} These consistent trends across model size and training type underscore that \textsc{RAGuard} probes weaknesses beyond those captured by conventional reasoning or robustness benchmarks.

\paragraph{Retrieval and Misleading Evidence.}\label{sec:retrievalandmisleading}
The fact that Oracle Retrieval leads to worse performance than Standard RAG suggests that retrieval errors, while suboptimal, may result in less damaging content than intentionally misleading documents, validating the construction and annotation of such documents in \textsc{RAGuard}.
To quantify this phenomenon, we introduce Misleading Retrieval Recall, which measures the proportion of claims for which at least one misleading document is retrieved. In RAG-1, this rate is 21.3\%, increasing to 44.8\% in RAG-5, indicating that retrieving more documents raises the likelihood of including harmful content. We find that when Misleading Retrieval Recall is higher, as in  Oracle Retrieval (Misleading), where Misleading Retrieval Recall is 100\%, the LLM performance decreases further, demonstrating the more damaging effect of \textsc{RAGuard}'s misleading documents. Additional retrieval metrics are reported in Appendix~\ref{sec:retrievalappendix}.

\paragraph{Qualitative Examples.}
Figure~\ref{fig:datasetexample} illustrates how misleading evidence interferes with LLM reasoning on specific claims. In the left example, the document provides an opinion questioning the claim without directly contradicting it (e.g., “less likely” does not invalidate the absoluteness of “anybody else”), yet the model misinterprets this subjective tone as factual evidence, a pattern we term confusing opinion with fact. 

In the right example, the misleading document introduces contradictory information, but the discrepancy can be resolved by temporal reasoning: The claim and the evidence refer to different time periods (“2010” versus “when COVID first began”). We observe that models frequently fail to make such contextual distinctions. By misapplying contextual cues, they overemphasize superficial signals like numbers or names while ignoring the broader meaning. These examples reveal how LLMs overly focus on surface-level indicators, explaining why they are highly susceptible to misleading retrievals, even when humans can resolve the ambiguity with relative ease. We provide additional examples of these failures in Appendix \ref{sec:qualitative}.



\section{Human Study}\label{sec:human-study}

To better understand how noisy context contained in \textsc{RAGuard} affect reasoning, 
we study human robustness to noisy contexts to compare 
to LLM performance.
We construct a 64-instance subset \begin{wrapfigure}{r}{0.5\linewidth}  
    \centering
        
    \includegraphics[width=0.98\linewidth]{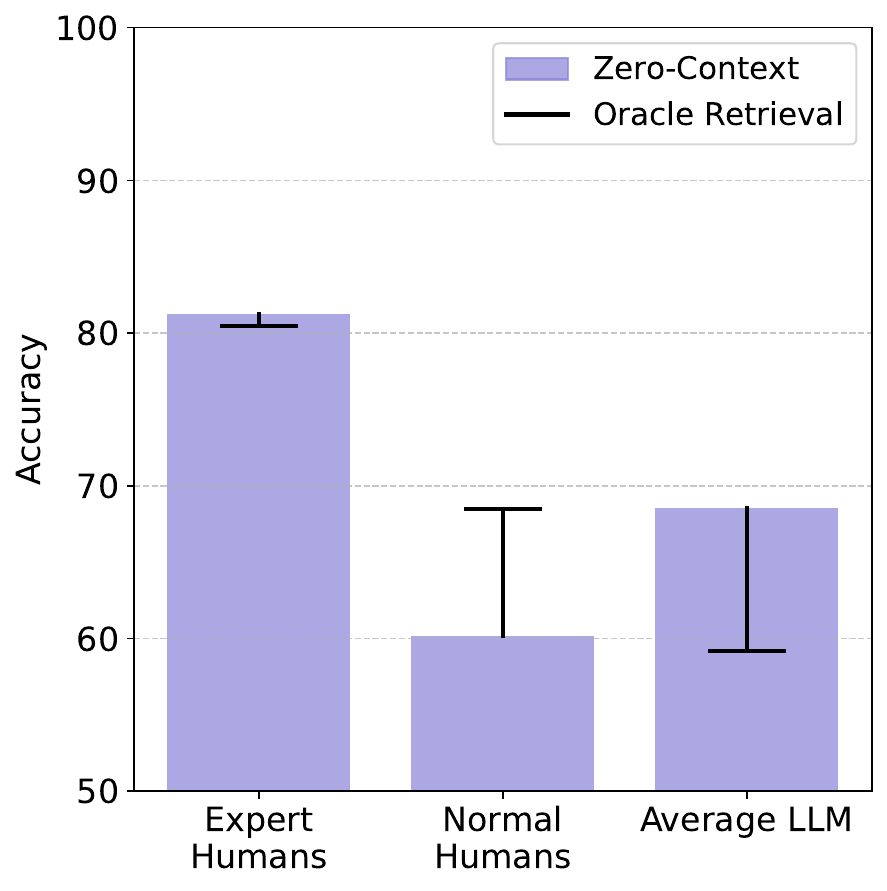}
    \caption{Performance of humans and LLMs on a subset of \textsc{RAGuard}, with black bars showing the performance differences when given documents.}
    \label{fig:human_results}
\end{wrapfigure}balanced across true and false claims, reflecting the distribution of misleading 
and supporting documents in our full dataset, while down-sampling unrelated documents. 
On this subset, all models perform within 5\% of the performance reported in Table~\ref{tab:main}. Of four recruited human annotators, two are PhD-level \textit{experts} in adjacent fields and two self-identify to know minimal information about politics beyond the basics (\textit{normal}). Annotator instructions are provided in Appendix~\ref{sec:humaninstruct}.


Shown in Figure~\ref{fig:human_results}, while normal humans benefit from having more information from a potentially misleading context, LLMs are easily misled by same information, illustrated by the longer black bar. In contrast, experts' consistently high accuracy suggests they do not benefit from additional information but can robustly handle misleading information. Rather than evaluating which of the three groups achieves higher accuracy on the zero-shot task, we focus on how they react to noisy context (i.e., the black bar in Figure~\ref{fig:human_results}). These results highlight a significant gap between human and machine reasoning.


\section{Limitations}\label{sec:limitationsmain}



While the approach of using GPT-4 for document annotation has a model-specific element, akin to how human annotation carries annotator perspectives, its utility lies in creating a testbed for a unique aspect of LLM robustness. We assessed bias through a cross-model agreement analysis, which found moderate to strong consistency across LLM annotators, and confirmed in Section \ref{sec:model-results} that all models, not simply GPT-4, fail similarly on the same examples. We further emphasize that our task is fact-checking based on gold verdicts from PolitiFact rather than misinformation detection. GPT-derived labels are used only for intermediate analysis (i.e., determination of which documents are included in the Oracle Retrieval (Misleading) setting).

\added{The limited scope of the human study constrains the strength of human–model comparisons, though it serves as a diagnostic check showing that observed failures arise from model reasoning rather than inherently impossible tasks. Ultimately, our main contribution lies in demonstrating how our dataset harms and misleads LLMs, as it is specifically constructed to challenge model reasoning. Whether these instances also mislead humans is not the primary focus. Scaling human evaluation for fact-checking remains an open problem that future work should address. }




\section{Conclusion}

In this paper, we highlight the importance of assessing the robustness of RAG systems against misleading retrievals, defining a new robustness-focused fact verification task that challenges models to
reason through misleading content while unifying inconsistent terminology in prior works into a structured framework. 
To advance the development of robust fact-checking systems, we introduce \textsc{RAGuard}, a diverse benchmark incorporating naturally occurring misleading data from Reddit discussions alongside verified evidence and claims from PolitiFact. Unlike prior RAG benchmarks that rely on synthetically noisy data, \textsc{RAGuard} utilizes real-world evidence, targeting cases where gold-standard documents may not exist. This mirrors the complexities of real-world misinformation, which is necessary for more robust systems. 

Our findings show that the performance of current RAG systems deteriorates significantly when exposed to misleading evidence,  challenging the assumption that retrieval always enhances model accuracy. Consequently, future research should focus on enhancing LLM robustness through methods such as adversarial retrieval training, which exposes models to misleading evidence during training to improve resilience. Additionally, incorporating multi-step reasoning and classifying documents for their subjectiveness can mitigate the impact of misleading sources. 

By providing a challenging yet realistic benchmark, \textsc{RAGuard} encourages the development of more sophisticated retrieval-based fact-checking methods. We hope this dataset will facilitate progress in designing retrieval pipelines that are not only effective but also resistant to misinformation, ultimately contributing to more reliable and trustworthy AI systems.

\bibliographystyle{plain}
\bibliography{neurips}

@article{schlichtkrull2023averitecdatasetrealworldclaim,
  title={Averitec: A dataset for real-world claim verification with evidence from the web},
  author={Schlichtkrull, Michael and Guo, Zhijiang and Vlachos, Andreas},
  journal={Advances in Neural Information Processing Systems},
  volume={36},
  pages={65128--65167},
  year={2023}
}

@inproceedings{yang2024rag,
  title={Im-rag: Multi-round retrieval-augmented generation through learning inner monologues},
  author={Yang, Diji and Rao, Jinmeng and Chen, Kezhen and Guo, Xiaoyuan and Zhang, Yawen and Yang, Jie and Zhang, Yi},
  booktitle={Proceedings of the 47th International ACM SIGIR Conference on Research and Development in Information Retrieval},
  pages={730--740},
  year={2024}
}

@article{tang2024minicheck,
  title={Minicheck: Efficient fact-checking of llms on grounding documents},
  author={Tang, Liyan and Laban, Philippe and Durrett, Greg},
  journal={arXiv preprint arXiv:2404.10774},
  year={2024}
}

@inproceedings{TriviaQA,
    title = "{T}rivia{QA}: A Large Scale Distantly Supervised Challenge Dataset for Reading Comprehension",
    author = "Joshi, Mandar  and
      Choi, Eunsol  and
      Weld, Daniel  and
      Zettlemoyer, Luke",
    editor = "Barzilay, Regina  and
      Kan, Min-Yen",
    booktitle = "Proceedings of the 55th Annual Meeting of the Association for Computational Linguistics (Volume 1: Long Papers)",
    month = jul,
    year = "2017",
    address = "Vancouver, Canada",
    publisher = "Association for Computational Linguistics",
    url = "https://aclanthology.org/P17-1147/",
    doi = "10.18653/v1/P17-1147",
    pages = "1601--1611"
}

@inproceedings{HotPotQA,
    title = "{H}otpot{QA}: A Dataset for Diverse, Explainable Multi-hop Question Answering",
    author = "Yang, Zhilin  and
      Qi, Peng  and
      Zhang, Saizheng  and
      Bengio, Yoshua  and
      Cohen, William  and
      Salakhutdinov, Ruslan  and
      Manning, Christopher D.",
    editor = "Riloff, Ellen  and
      Chiang, David  and
      Hockenmaier, Julia  and
      Tsujii, Jun{'}ichi",
    booktitle = "Proceedings of the 2018 Conference on Empirical Methods in Natural Language Processing",
    month = oct # "-" # nov,
    year = "2018",
    address = "Brussels, Belgium",
    publisher = "Association for Computational Linguistics",
    url = "https://aclanthology.org/D18-1259/",
    doi = "10.18653/v1/D18-1259",
    pages = "2369--2380"
}

@inproceedings{SQUAD,
    title = "{SQ}u{AD}: 100,000+ Questions for Machine Comprehension of Text",
    author = "Rajpurkar, Pranav  and
      Zhang, Jian  and
      Lopyrev, Konstantin  and
      Liang, Percy",
    editor = "Su, Jian  and
      Duh, Kevin  and
      Carreras, Xavier",
    booktitle = "Proceedings of the 2016 Conference on Empirical Methods in Natural Language Processing",
    month = nov,
    year = "2016",
    address = "Austin, Texas",
    publisher = "Association for Computational Linguistics",
    url = "https://aclanthology.org/D16-1264/",
    doi = "10.18653/v1/D16-1264",
    pages = "2383--2392"
}

@article{NaturalQuestions,
    title = "Natural Questions: A Benchmark for Question Answering Research",
    author = "Kwiatkowski, Tom  and
      Palomaki, Jennimaria  and
      Redfield, Olivia  and
      Collins, Michael  and
      Parikh, Ankur  and
      Alberti, Chris  and
      Epstein, Danielle  and
      Polosukhin, Illia  and
      Devlin, Jacob  and
      Lee, Kenton  and
      Toutanova, Kristina  and
      Jones, Llion  and
      Kelcey, Matthew  and
      Chang, Ming-Wei  and
      Dai, Andrew M.  and
      Uszkoreit, Jakob  and
      Le, Quoc  and
      Petrov, Slav",
    editor = "Lee, Lillian  and
      Johnson, Mark  and
      Roark, Brian  and
      Nenkova, Ani",
    journal = "Transactions of the Association for Computational Linguistics",
    volume = "7",
    year = "2019",
    address = "Cambridge, MA",
    publisher = "MIT Press",
    url = "https://aclanthology.org/Q19-1026",
    doi = "10.1162/tacl_a_00276",
    pages = "452--466",
}

@article{PolitifactOslo,
  title={The PolitiFact-Oslo Corpus: A new dataset for fake news analysis and detection},
  author={P{\~o}ldvere, Nele and Uddin, Zia and Thomas, Aleena},
  journal={Information},
  volume={14},
  number={12},
  pages={627},
  year={2023},
  publisher={MDPI}
}

@inproceedings{Liar,
    title = "{``}Liar, Liar Pants on Fire{''}: A New Benchmark Dataset for Fake News Detection",
    author = "Wang, William Yang",
    editor = "Barzilay, Regina  and
      Kan, Min-Yen",
    booktitle = "Proceedings of the 55th Annual Meeting of the Association for Computational Linguistics (Volume 2: Short Papers)",
    month = jul,
    year = "2017",
    address = "Vancouver, Canada",
    publisher = "Association for Computational Linguistics",
    url = "https://aclanthology.org/P17-2067",
    doi = "10.18653/v1/P17-2067",
    pages = "422--426",
}

@inproceedings{FEVER,
    title = "{FEVER}: a Large-scale Dataset for Fact Extraction and {VER}ification",
    author = "Thorne, James  and
      Vlachos, Andreas  and
      Christodoulopoulos, Christos  and
      Mittal, Arpit",
    editor = "Walker, Marilyn  and
      Ji, Heng  and
      Stent, Amanda",
    booktitle = "Proceedings of the 2018 Conference of the North {A}merican Chapter of the Association for Computational Linguistics: Human Language Technologies, Volume 1 (Long Papers)",
    month = jun,
    year = "2018",
    address = "New Orleans, Louisiana",
    publisher = "Association for Computational Linguistics",
    url = "https://aclanthology.org/N18-1074",
    doi = "10.18653/v1/N18-1074",
    pages = "809--819",
}

@inproceedings{PowerOfNoise, series={SIGIR 2024},
   title={The Power of Noise: Redefining Retrieval for RAG Systems},
   url={http://dx.doi.org/10.1145/3626772.3657834},
   DOI={10.1145/3626772.3657834},
   booktitle={Proceedings of the 47th International ACM SIGIR Conference on Research and Development in Information Retrieval},
   publisher={ACM},
   author={Cuconasu, Florin and Trappolini, Giovanni and Siciliano, Federico and Filice, Simone and Campagnano, Cesare and Maarek, Yoelle and Tonellotto, Nicola and Silvestri, Fabrizio},
   year={2024},
   month=jul, pages={719–729},
   collection={SIGIR 2024} }

@article{openai2024gpt4technicalreport,
  title={GPT-4 technical report},
  author={Achiam, Josh and Adler, Steven and Agarwal, Sandhini and Ahmad, Lama and Akkaya, Ilge and Aleman, Florencia Leoni and Almeida, Diogo and Altenschmidt, Janko and Altman, Sam and Anadkat, Shyamal and others},
  journal={arXiv preprint arXiv:2303.08774},
  year={2023}
}

@article{AstuteRAG,
  title={Astute rag: Overcoming imperfect retrieval augmentation and knowledge conflicts for large language models},
  author={Wang, Fei and Wan, Xingchen and Sun, Ruoxi and Chen, Jiefeng and Ar{\i}k, Sercan {\"O}},
  journal={arXiv preprint arXiv:2410.07176},
  year={2024}
}

@misc{openai2025o4mini,
  title        = {OpenAI o3 and o4-mini System Card},
  author       = {{OpenAI}},
  year         = {2025},
  month        = {April},
  url          = {https://cdn.openai.com/pdf/2221c875-02dc-4789-800b-e7758f3722c1/o3-and-o4-mini-system-card.pdf},
  note         = {Accessed: 2025-05-10}
}

@article{deepseekai2025deepseekr1incentivizingreasoningcapability,
  title={Deepseek-r1: Incentivizing reasoning capability in llms via reinforcement learning},
  author={Guo, Daya and Yang, Dejian and Zhang, Haowei and Song, Junxiao and Zhang, Ruoyu and Xu, Runxin and Zhu, Qihao and Ma, Shirong and Wang, Peiyi and Bi, Xiao and others},
  journal={arXiv preprint arXiv:2501.12948},
  year={2025}
}

@article{ContextMemoryConflict,
  title={Studying Large Language Model Behaviors Under Context-Memory Conflicts With Real Documents},
  author={Kortukov, Evgenii and Rubinstein, Alexander and Nguyen, Elisa and Oh, Seong Joon},
  journal={arXiv preprint arXiv:2404.16032},
  year={2024}
}

@inproceedings{xiong2024benchmarkingretrievalaugmentedgenerationmedicine,
    title = "Benchmarking Retrieval-Augmented Generation for Medicine",
    author = "Xiong, Guangzhi  and
      Jin, Qiao  and
      Lu, Zhiyong  and
      Zhang, Aidong",
    editor = "Ku, Lun-Wei  and
      Martins, Andre  and
      Srikumar, Vivek",
    booktitle = "Findings of the Association for Computational Linguistics: ACL 2024",
    month = aug,
    year = "2024",
    address = "Bangkok, Thailand",
    publisher = "Association for Computational Linguistics",
    url = "https://aclanthology.org/2024.findings-acl.372/",
    doi = "10.18653/v1/2024.findings-acl.372",
    pages = "6233--6251"
}

@article{guha2024legalbench,
  title={Legalbench: A collaboratively built benchmark for measuring legal reasoning in large language models},
  author={Guha, Neel and Nyarko, Julian and Ho, Daniel and R{\'e}, Christopher and Chilton, Adam and Chohlas-Wood, Alex and Peters, Austin and Waldon, Brandon and Rockmore, Daniel and Zambrano, Diego and others},
  journal={Advances in Neural Information Processing Systems},
  volume={36},
  pages={44123--44279},
  year={2023}
}

@InProceedings{RAGPretrain,
  title = 	 {Retrieval Augmented Language Model Pre-Training},
  author =       {Guu, Kelvin and Lee, Kenton and Tung, Zora and Pasupat, Panupong and Chang, Mingwei},
  booktitle = 	 {Proceedings of the 37th International Conference on Machine Learning},
  pages = 	 {3929--3938},
  year = 	 {2020},
  editor = 	 {III, Hal Daumé and Singh, Aarti},
  volume = 	 {119},
  series = 	 {Proceedings of Machine Learning Research},
  month = 	 {13--18 Jul},
  publisher =    {PMLR},
  url = 	 {https://proceedings.mlr.press/v119/guu20a.html}
}

@article{RAGNLP,
  title={Retrieval-augmented generation for knowledge-intensive nlp tasks},
  author={Lewis, Patrick and Perez, Ethan and Piktus, Aleksandra and Petroni, Fabio and Karpukhin, Vladimir and Goyal, Naman and K{\"u}ttler, Heinrich and Lewis, Mike and Yih, Wen-tau and Rockt{\"a}schel, Tim and others},
  journal={Advances in neural information processing systems},
  volume={33},
  pages={9459--9474},
  year={2020}
}

@inproceedings{xie2024adaptivechameleonstubbornsloth,
  title={Adaptive chameleon or stubborn sloth: Revealing the behavior of large language models in knowledge conflicts},
  author={Xie, Jian and Zhang, Kai and Chen, Jiangjie and Lou, Renze and Su, Yu},
  booktitle={The Twelfth International Conference on Learning Representations},
  year={2023}
}

@inproceedings{feverous,
    title = "The Fact Extraction and {VER}ification Over Unstructured and Structured information ({FEVEROUS}) Shared Task",
    author = "Aly, Rami  and
      Guo, Zhijiang  and
      Schlichtkrull, Michael Sejr  and
      Thorne, James  and
      Vlachos, Andreas  and
      Christodoulopoulos, Christos  and
      Cocarascu, Oana  and
      Mittal, Arpit",
    editor = "Aly, Rami  and
      Christodoulopoulos, Christos  and
      Cocarascu, Oana  and
      Guo, Zhijiang  and
      Mittal, Arpit  and
      Schlichtkrull, Michael  and
      Thorne, James  and
      Vlachos, Andreas",
    booktitle = "Proceedings of the Fourth Workshop on Fact Extraction and VERification (FEVER)",
    month = nov,
    year = "2021",
    address = "Dominican Republic",
    publisher = "Association for Computational Linguistics",
    url = "https://aclanthology.org/2021.fever-1.1/",
    doi = "10.18653/v1/2021.fever-1.1",
    pages = "1--13"
}

@inproceedings{thorne-vlachos-2018-automated,
    title = "Automated Fact Checking: Task Formulations, Methods and Future Directions",
    author = "Thorne, James  and
      Vlachos, Andreas",
    editor = "Bender, Emily M.  and
      Derczynski, Leon  and
      Isabelle, Pierre",
    booktitle = "Proceedings of the 27th International Conference on Computational Linguistics",
    month = aug,
    year = "2018",
    address = "Santa Fe, New Mexico, USA",
    publisher = "Association for Computational Linguistics",
    url = "https://aclanthology.org/C18-1283/",
    pages = "3346--3359"
}

@inproceedings{borgeaud2022improving,
  title={Improving language models by retrieving from trillions of tokens},
  author={Borgeaud, Sebastian and Mensch, Arthur and Hoffmann, Jordan and Cai, Trevor and Rutherford, Eliza and Millican, Katie and Van Den Driessche, George Bm and Lespiau, Jean-Baptiste and Damoc, Bogdan and Clark, Aidan and others},
  booktitle={International conference on machine learning},
  pages={2206--2240},
  year={2022},
  organization={PMLR}
}

@inproceedings{karpukhin-etal-2020-dense,
    title = "Dense Passage Retrieval for Open-Domain Question Answering",
    author = "Karpukhin, Vladimir  and
      Oguz, Barlas  and
      Min, Sewon  and
      Lewis, Patrick  and
      Wu, Ledell  and
      Edunov, Sergey  and
      Chen, Danqi  and
      Yih, Wen-tau",
    editor = "Webber, Bonnie  and
      Cohn, Trevor  and
      He, Yulan  and
      Liu, Yang",
    booktitle = "Proceedings of the 2020 Conference on Empirical Methods in Natural Language Processing (EMNLP)",
    month = nov,
    year = "2020",
    address = "Online",
    publisher = "Association for Computational Linguistics",
    url = "https://aclanthology.org/2020.emnlp-main.550/",
    doi = "10.18653/v1/2020.emnlp-main.550",
    pages = "6769--6781"
}

@article{RobustRAG,
  title={Certifiably robust rag against retrieval corruption},
  author={Xiang, Chong and Wu, Tong and Zhong, Zexuan and Wagner, David and Chen, Danqi and Mittal, Prateek},
  journal={arXiv preprint arXiv:2405.15556},
  year={2024}
}

@article{InstructRAG,
  title={InstructRAG: Instructing Retrieval-Augmented Generation via Self-Synthesized Rationales},
  author={Wei, Zhepei and Chen, Wei-Lin and Meng, Yu},
  journal={arXiv preprint arXiv:2406.13629},
  year={2024}
}

@article{LearningToBreak,
  title={Learning to break: Knowledge-enhanced reasoning in multi-agent debate system},
  author={Wang, Haotian and Du, Xiyuan and Yu, Weijiang and Chen, Qianglong and Zhu, Kun and Chu, Zheng and Yan, Lian and Guan, Yi},
  journal={Neurocomputing},
  volume={618},
  pages={129063},
  year={2025},
  publisher={Elsevier}
}

@inproceedings{mocheg,
  title={End-to-end multimodal fact-checking and explanation generation: A challenging dataset and models},
  author={Yao, Barry Menglong and Shah, Aditya and Sun, Lichao and Cho, Jin-Hee and Huang, Lifu},
  booktitle={Proceedings of the 46th International ACM SIGIR Conference on Research and Development in Information Retrieval},
  pages={2733--2743},
  year={2023}
}

@article{NoiserBench,
  title={Pandora's Box or Aladdin's Lamp: A Comprehensive Analysis Revealing the Role of RAG Noise in Large Language Models},
  author={Wu, Jinyang and Zhang, Shuai and Che, Feihu and Feng, Mingkuan and Zhang, Chuyuan and Shao, Pengpeng and Tao, Jianhua},
  journal={arXiv preprint arXiv:2408.13533},
  year={2024}
}

@article{FakeNewsNet,
  title={Fakenewsnet: A data repository with news content, social context, and spatiotemporal information for studying fake news on social media},
  author={Shu, Kai and Mahudeswaran, Deepak and Wang, Suhang and Lee, Dongwon and Liu, Huan},
  journal={Big data},
  volume={8},
  number={3},
  pages={171--188},
  year={2020},
  publisher={Mary Ann Liebert, Inc., publishers 140 Huguenot Street, 3rd Floor New~…}
}

@inproceedings{nielsen2022mumin,
  title={Mumin: A large-scale multilingual multimodal fact-checked misinformation social network dataset},
  author={Nielsen, Dan S and McConville, Ryan},
  booktitle={Proceedings of the 45th international ACM SIGIR conference on research and development in information retrieval},
  pages={3141--3153},
  year={2022}
}

@inproceedings{snopes,
    title = "A Richly Annotated Corpus for Different Tasks in Automated Fact-Checking",
    author = "Hanselowski, Andreas  and
      Stab, Christian  and
      Schulz, Claudia  and
      Li, Zile  and
      Gurevych, Iryna",
    editor = "Bansal, Mohit  and
      Villavicencio, Aline",
    booktitle = "Proceedings of the 23rd Conference on Computational Natural Language Learning (CoNLL)",
    month = nov,
    year = "2019",
    address = "Hong Kong, China",
    publisher = "Association for Computational Linguistics",
    url = "https://aclanthology.org/K19-1046/",
    doi = "10.18653/v1/K19-1046",
    pages = "493--503"
}

@article{dubey2024llama,
  title={The llama 3 herd of models},
  author={Dubey, Abhimanyu and Jauhri, Abhinav and Pandey, Abhinav and Kadian, Abhishek and Al-Dahle, Ahmad and Letman, Aiesha and Mathur, Akhil and Schelten, Alan and Yang, Amy and Fan, Angela and others},
  journal={arXiv preprint arXiv:2407.21783},
  year={2024}
}

@article{jiang2023mistral7b,
  title={Mistral 7b},
  author={Jiang, Albert Q and Sablayrolles, A and Mensch, A and Bamford, C and Chaplot, D Singh and Casas, Ddl and Bressand, F and Lengyel, G and Lample, G and Saulnier, L and others},
  journal={arXiv preprint arXiv:2310.06825},
  year={2023}
}

@article{gao2023retrieval,
  title={Retrieval-augmented generation for large language models: A survey},
  author={Gao, Yunfan and Xiong, Yun and Gao, Xinyu and Jia, Kangxiang and Pan, Jinliu and Bi, Yuxi and Dai, Yi and Sun, Jiawei and Wang, Haofen},
  journal={arXiv preprint arXiv:2312.10997},
  year={2023}
}

@article{geminiteam2024gemini15unlockingmultimodal,
  title={Gemini 1.5: Unlocking multimodal understanding across millions of tokens of context},
  author={Team, Gemini and Georgiev, Petko and Lei, Ving Ian and Burnell, Ryan and Bai, Libin and Gulati, Anmol and Tanzer, Garrett and Vincent, Damien and Pan, Zhufeng and Wang, Shibo and others},
  journal={arXiv preprint arXiv:2403.05530},
  year={2024}
}

@misc{claude2024,
	title={Claude 3.5 Sonnet},
	url={https://www.anthropic.com/news/claude-3-5-sonnet},
	author={Anthropic},
	month={June},
	year={2024},
    note         = {Accessed: 2025-05-10}
}

@inproceedings{pubhealth,
    title = "Explainable Automated Fact-Checking for Public Health Claims",
    author = "Kotonya, Neema  and
      Toni, Francesca",
    editor = "Webber, Bonnie  and
      Cohn, Trevor  and
      He, Yulan  and
      Liu, Yang",
    booktitle = "Proceedings of the 2020 Conference on Empirical Methods in Natural Language Processing (EMNLP)",
    month = nov,
    year = "2020",
    address = "Online",
    publisher = "Association for Computational Linguistics",
    url = "https://aclanthology.org/2020.emnlp-main.623/",
    doi = "10.18653/v1/2020.emnlp-main.623",
    pages = "7740--7754"
}

@inproceedings{multifc,
    title = "{M}ulti{FC}: A Real-World Multi-Domain Dataset for Evidence-Based Fact Checking of Claims",
    author = "Augenstein, Isabelle  and
      Lioma, Christina  and
      Wang, Dongsheng  and
      Chaves Lima, Lucas  and
      Hansen, Casper  and
      Hansen, Christian  and
      Simonsen, Jakob Grue",
    editor = "Inui, Kentaro  and
      Jiang, Jing  and
      Ng, Vincent  and
      Wan, Xiaojun",
    booktitle = "Proceedings of the 2019 Conference on Empirical Methods in Natural Language Processing and the 9th International Joint Conference on Natural Language Processing (EMNLP-IJCNLP)",
    month = nov,
    year = "2019",
    address = "Hong Kong, China",
    publisher = "Association for Computational Linguistics",
    url = "https://aclanthology.org/D19-1475/",
    doi = "10.18653/v1/D19-1475",
    pages = "4685--4697"
}

@inproceedings{qacc,
    title = "Open Domain Question Answering with Conflicting Contexts",
    author = "Liu, Siyi  and
      Ning, Qiang  and
      Halder, Kishaloy  and
      Qi, Zheng  and
      Xiao, Wei  and
      Htut, Phu Mon  and
      Zhang, Yi  and
      Anna John, Neha  and
      Min, Bonan  and
      Benajiba, Yassine  and
      Roth, Dan",
    editor = "Chiruzzo, Luis  and
      Ritter, Alan  and
      Wang, Lu",
    booktitle = "Findings of the Association for Computational Linguistics: NAACL 2025",
    month = apr,
    year = "2025",
    address = "Albuquerque, New Mexico",
    publisher = "Association for Computational Linguistics",
    url = "https://aclanthology.org/2025.findings-naacl.99/",
    doi = "10.18653/v1/2025.findings-naacl.99",
    pages = "1838--1854",
    ISBN = "979-8-89176-195-7"
}

@inproceedings{raat,
    title = "Enhancing Noise Robustness of Retrieval-Augmented Language Models with Adaptive Adversarial Training",
    author = "Fang, Feiteng  and
      Bai, Yuelin  and
      Ni, Shiwen  and
      Yang, Min  and
      Chen, Xiaojun  and
      Xu, Ruifeng",
    editor = "Ku, Lun-Wei  and
      Martins, Andre  and
      Srikumar, Vivek",
    booktitle = "Proceedings of the 62nd Annual Meeting of the Association for Computational Linguistics (Volume 1: Long Papers)",
    month = aug,
    year = "2024",
    address = "Bangkok, Thailand",
    publisher = "Association for Computational Linguistics",
    url = "https://aclanthology.org/2024.acl-long.540/",
    doi = "10.18653/v1/2024.acl-long.540",
    pages = "10028--10039"
}

@inproceedings{yin2023alcunalargelanguagemodels,
    title = "{ALCUNA}: Large Language Models Meet New Knowledge",
    author = "Yin, Xunjian  and
      Huang, Baizhou  and
      Wan, Xiaojun",
    editor = "Bouamor, Houda  and
      Pino, Juan  and
      Bali, Kalika",
    booktitle = "Proceedings of the 2023 Conference on Empirical Methods in Natural Language Processing",
    month = dec,
    year = "2023",
    address = "Singapore",
    publisher = "Association for Computational Linguistics",
    url = "https://aclanthology.org/2023.emnlp-main.87/",
    doi = "10.18653/v1/2023.emnlp-main.87",
    pages = "1397--1414"
}

@inproceedings{relevance,
author = {Sauchuk, Artsiom and Thorne, James and Halevy, Alon and Tonellotto, Nicola and Silvestri, Fabrizio},
title = {On the Role of Relevance in Natural Language Processing Tasks},
year = {2022},
isbn = {9781450387323},
publisher = {Association for Computing Machinery},
address = {New York, NY, USA},
url = {https://doi.org/10.1145/3477495.3532034},
doi = {10.1145/3477495.3532034},
booktitle = {Proceedings of the 45th International ACM SIGIR Conference on Research and Development in Information Retrieval},
pages = {1785–1789},
numpages = {5},
location = {Madrid, Spain},
series = {SIGIR '22}
}

@article{groeneveld2024olmo,
  title={Olmo: Accelerating the science of language models},
  author={Groeneveld, Dirk and Beltagy, Iz and Walsh, Pete and Bhagia, Akshita and Kinney, Rodney and Tafjord, Oyvind and Jha, Ananya Harsh and Ivison, Hamish and Magnusson, Ian and Wang, Yizhong and others},
  journal={arXiv preprint arXiv:2402.00838},
  year={2024}
}

@article{yan2024corrective,
  title={Corrective Retrieval Augmented Generation},
  author={Yan, Shi-Qi and Gu, Jia-Chen and Zhu, Yun and Ling, Zhen-Hua},
  journal={arXiv preprint arXiv:2401.15884},
  year={2024}
}

\newpage

\begin{enumerate}

\item {\bf Claims}
    \item[] Question: Do the main claims made in the abstract and introduction accurately reflect the paper's contributions and scope?
    \item[] Answer: \answerYes{} 
    \item[] Justification: The abstract and introduction accurately describe the problem (RAG robustness), proposed benchmark (\textsc{RAGuard}), and experimental scope (fact-checking with supporting, misleading, and unrelated evidence from Reddit), which align with the content presented in the paper (Sections 4, 5, Appendix D, E).


\item {\bf Limitations}
    \item[] Question: Does the paper discuss the limitations of the work performed by the authors?
    \item[] Answer: \answerYes{} 
    \item[] Justification: Section 6 discusses limitations, including the use of GPT-4 for annotation, the distribution of document labels, and the size of the human study.

\item {\bf Theory assumptions and proofs}
    \item[] Question: For each theoretical result, does the paper provide the full set of assumptions and a complete (and correct) proof?
    \item[] Answer: \answerNA{} 
    \item[] Justification: 

    \item {\bf Experimental result reproducibility}
    \item[] Question: Does the paper fully disclose all the information needed to reproduce the main experimental results of the paper to the extent that it affects the main claims and/or conclusions of the paper (regardless of whether the code and data are provided or not)?
    \item[] Answer: \answerYes{} 
    \item[] Justification: The paper describes the data construction method (Section 3) and experimental setup (Section 4). The appendix provides hyperparameter details and prompts for data construction and baseline evaluation (Appendix C). 

\item {\bf Open access to data and code}
    \item[] Question: Does the paper provide open access to the data and code, with sufficient instructions to faithfully reproduce the main experimental results, as described in supplemental material?
    \item[] Answer: \answerYes{} 
    \item[] Justification: The dataset is provided. The paper focuses on introducing the dataset and uses straightforward LLM calls for baselines. All methods are reproducible through the information provided in the paper.

\item {\bf Experimental setting/details}
    \item[] Question: Does the paper specify all the training and test details (e.g., data splits, hyperparameters, how they were chosen, type of optimizer, etc.) necessary to understand the results?
    \item[] Answer: \answerYes{} 
    \item[] Justification: Appendix C provides hyperparameter details and prompts for data construction and baseline evaluation. The paper describes the data construction method (Section 3) and experimental setup (Section 4).  

\item {\bf Experiment statistical significance}
    \item[] Question: Does the paper report error bars suitably and correctly defined or other appropriate information about the statistical significance of the experiments?
    \item[] Answer: \answerNo{} 
    \item[] Justification: We mitigate randomness by evaluating multiple models with different architectures and consistently observe substantial accuracy drops across all models and settings (Section 4), indicating a clear and reproducible performance degradation trend.

\item {\bf Experiments compute resources}
    \item[] Question: For each experiment, does the paper provide sufficient information on the computer resources (type of compute workers, memory, time of execution) needed to reproduce the experiments?
    \item[] Answer: \answerYes{} 
    \item[] Justification: The compute resources required are bounded by the LLM. We do not add any complexity beyond using each selected LLM for inference. Refer to LLM’s original paper for precise requirements.
    
\item {\bf Code of ethics}
    \item[] Question: Does the research conducted in the paper conform, in every respect, with the NeurIPS Code of Ethics \url{https://neurips.cc/public/EthicsGuidelines}?
    \item[] Answer: \answerYes{} 
    \item[] Justification: Our work involves only publicly available, de-identified data and does not include human subjects or sensitive personal information. The dataset does not include protected categories and is not intended for high-risk applications. We also clearly document the dataset’s scope, limitations, and intended use (in the publicly available repository) to mitigate any potential downstream harm.

\item {\bf Broader impacts}
    \item[] Question: Does the paper discuss both potential positive societal impacts and negative societal impacts of the work performed?
    \item[] Answer: \answerYes{} 
    \item[] Justification: The societal impacts are discussed in Section 1 and 7. Negative impacts are elaborated in the dataset repository: This dataset has been compiled from publicly available sources on the Internet. It may contain discussions on sensitive political topics, including viewpoints that some individuals may find controversial or offensive. The inclusion of any content does not imply endorsement of any views expressed. Users are advised to exercise discretion and ensure compliance with applicable ethical guidelines and legal frameworks when using this dataset.
    
\item {\bf Safeguards}
    \item[] Question: Does the paper describe safeguards that have been put in place for responsible release of data or models that have a high risk for misuse (e.g., pretrained language models, image generators, or scraped datasets)?
    \item[] Answer: \answerYes{} 
    \item[] Justification: Safeguards are elaborated in the dataset repository: This dataset has been compiled from publicly available sources on the Internet. It may contain discussions on sensitive political topics, including viewpoints that some individuals may find controversial or offensive. The inclusion of any content does not imply endorsement of any views expressed. Users are advised to exercise discretion and ensure compliance with applicable ethical guidelines and legal frameworks when using this dataset.

\item {\bf Licenses for existing assets}
    \item[] Question: Are the creators or original owners of assets (e.g., code, data, models), used in the paper, properly credited and are the license and terms of use explicitly mentioned and properly respected?
    \item[] Answer: \answerYes{} 
    \item[] Justification: The dataset was created from publically available sources (Reddit and PolitiFact), which were credited. All search engines and LLMs used were properly cited.


\item {\bf New assets}
    \item[] Question: Are new assets introduced in the paper well documented and is the documentation provided alongside the assets?
    \item[] Answer: \answerYes{} 
    \item[] Justification: We follow the instructions from the NeurIPS single-blind submission when providing the new dataset.

\item {\bf Crowdsourcing and research with human subjects}
    \item[] Question: For crowdsourcing experiments and research with human subjects, does the paper include the full text of instructions given to participants and screenshots, if applicable, as well as details about compensation (if any)? 
    \item[] Answer: \answerYes{} 
    \item[] Justification: A small number of research team members helped verify a subset of the generated data. They received clear written instructions, and we include these instructions in the supplementary material. No monetary compensation was provided. Participants will be acknowledged in the final version.

\item {\bf Institutional review board (IRB) approvals or equivalent for research with human subjects}
    \item[] Question: Does the paper describe potential risks incurred by study participants, whether such risks were disclosed to the subjects, and whether Institutional Review Board (IRB) approvals (or an equivalent approval/review based on the requirements of your country or institution) were obtained?
    \item[] Answer: \answerNA{} 
    \item[] Justification: The task posed minimal risk and was conducted by internal research team members who will be acknowledged. Due to the minimal nature of the activity and low risk, no IRB approval was sought. Procedures followed our institution’s standard ethical guidelines.

\item {\bf Declaration of LLM usage}
    \item[] Question: Does the paper describe the usage of LLMs if it is an important, original, or non-standard component of the core methods in this research? Note that if the LLM is used only for writing, editing, or formatting purposes and does not impact the core methodology, scientific rigorousness, or originality of the research, declaration is not required.
    \item[] Answer: \answerYes{} 
    \item[] Justification: This paper tests the robustness of LLMs to misleading retrievals. Consequently, we use LLM performance as a way to categorize documents for their ability to mislead an LLM. There, we benchmark many commonly-used LLMs on this dataset to highlight the gaps in their robustness. Notably, other AI models can also be used in a similar way to create or test this dataset, but we focus on LLMs due to their relevance and widespread use, particularly in RAG.

\end{enumerate}

\newpage
\appendix

\section{Definitions} \label{sec:terminology}

Prior works employ varying terminology to describe the presence of such noise in retrieved contexts or retrieval corpora \cite{PowerOfNoise,qacc,raat,NoiserBench}. To establish consistency, we define a structured taxonomy and align existing definitions (See Figure~\ref{fig:definition}).

Typical RAG datasets, including all prior fact-checking datasets to our knowledge, exclusively contain non-noisy, supporting documents as associated evidence, leading to overly optimistic performance \cite{PowerOfNoise}. 
Instead of relying solely on answer-containing documents, our dataset adopts a broader notion of supporting evidence.
Specifically, we consider a document to be supporting if it provides information that enables an LLM to infer the correct answer, even if it does not explicitly state the ground-truth output. This reflects real-world fact-checking, where human verifiers rely on contextual information rather than single authoritative documents.


We categorize different types of noisy evidence based on whether the information directly conflicts with aspects of the correct prediction. As in prior work \cite{PowerOfNoise,raat}, we include non-conflicting documents in \textsc{RAGuard}, such as unrelated texts that may hurt performance. However, our primary focus is conflicting documents, which include misleading, fabricated, and unambiguous evidence. 
Previous datasets primarily include conflicting evidence as fabricated or unambiguous documents, oversimplifying real-world complexity and ambiguity (see Section \ref{sec:comparison} for further discussion) \cite{NoiserBench,qacc,raat}. Notably, no prior work has introduced misleading documents.

In \textsc{RAGuard}, misleading documents distort facts through selective framing, omission, or biased presentation, leading the system toward incorrect predictions while still containing partial truths. Unlike fabricated evidence, which is explicitly engineered to contradict the correct prediction (i.e., adversarial perturbations), misleading evidence subtly misguides the model rather than directly opposing it. Additionally, while prior work such as QACC \cite{qacc} introduces unambiguous evidence—a term we adopt to ensure consistency with past research—which includes some naturally conflicting evidence but only for a limited set of unambiguous questions, we focus on more natural yet scalable conflicting evidence.

For reference, we provide a list of all defined terms. Each term defines a type of document or piece of evidence.

\begin{enumerate}
    \item \textit{Associated:} any document linked to a claim, regardless of label

  \item \textit{Supporting:}
  aids the system in producing a correct prediction through containing the correct answer explicitly or providing contextual support 
  \item \textit{Noisy:} challenges or disrupt system performance, thereby enhancing robustness
 \item \textit{Conflicting:} contradicts either the correct answer or some aspect of the prediction
 \item \textit{Misleading:}  introduces factual distortions through selective framing, omission, or biased presentation; may contain partial truths
 \item \textit{Fabricated:} synthetically constructed to include factual errors (e.g., adversarial perturbations)
  \item \textit{Unambiguous:} naturally conflicting evidence but only for a limited set of unambiguous questions (special case of \cite{qacc})
\item \textit{Non-Conflicting:} does not directly contradict the correct answer but still introduces noise by distracting the model
\item \textit{Unrelated:} does not contain specific enough information to determine the correct prediction, despite being topically or semantically related to the query
\item \textit{Random:} unrelated; often introduced through random selection or artificial generation
  
\end{enumerate}

\section{Dataset Statistics}\label{sec:stats}

Note that \textsc{RAGuard} includes many unrelated documents, reflecting the natural distribution of the internet, where many discussions are neutral, neither misleading nor supporting the validity of a claim. These unrelated documents are topically relevant to their corresponding claim as they are retrieved with keyword search in the Dataset Construction stage (Section~\ref{sec:knowledge}). However, they are labeled as \textit{unrelated} because they do not specifically provide misleading or supporting content (Section~\ref{sec:llm-annotation}). \added{Although less directly misleading, these documents still significantly reduce LLM accuracy (e.g., Llama 3 drops from 61.1\% to 50.2\%).}

\begin{figure}[ht]
\centering

\begin{subfigure}{0.65\linewidth}
    \centering
    \includegraphics[width=\linewidth]{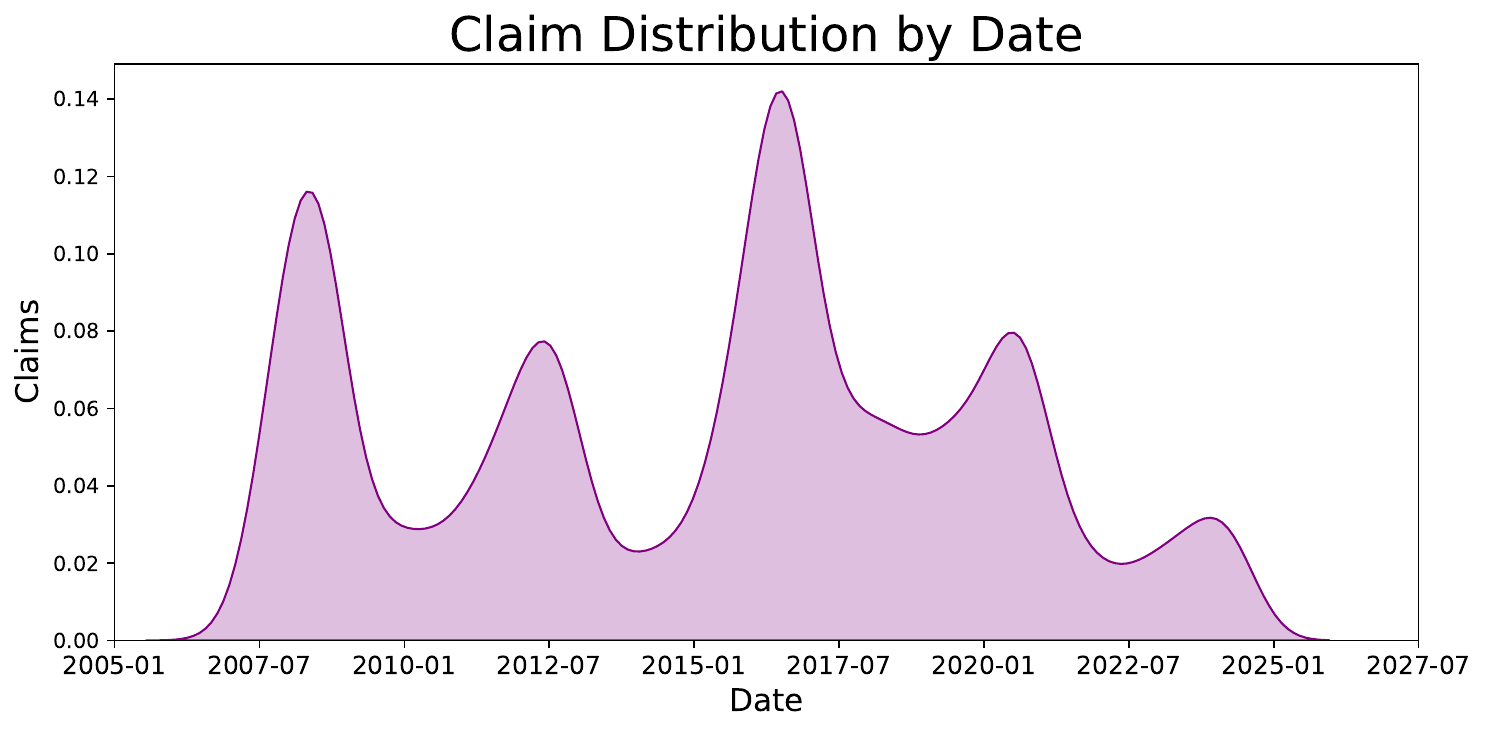}
\end{subfigure}%
\hfill
\begin{subfigure}{0.35\linewidth}
    \centering
    \resizebox{!}{0.4\linewidth}{
    \begin{tabular}{l r}
    \hline
    \textbf{Speaker} & \textbf{Claims} \\
    \hline
    Al Gore         & 6   \\
    George W. Bush  & 4   \\
    John Kerry      & 14  \\
    John McCain     & 173 \\
    Barack Obama    & 589 \\
    Mitt Romney     & 204 \\
    Hillary Clinton & 295 \\
    Donald Trump    & 1008 \\
    Joe Biden       & 308 \\
    Kamala Harris   & 47  \\
    \hline
    \end{tabular}}
    \vspace{0.5cm}
\end{subfigure}

\caption{Claim distribution by date and speaker.}
\label{fig:statistics}

\end{figure}

We provide additional dataset statistics in Figure~\ref{fig:statistics}. We see a spike in claims during recent presidential elections (2012, 2016, and 2024) which is an expected time for increased political discussion. We also note that more recent candidates usually have more claims in our dataset. This reflects the increasing popularity of fact-checking in recent years.

\section{Prompt Templates} \label{sec:prompts}

For all settings and models, we used a temperature of 0.1.

\subsection{Dataset Creation}\label{sec:prompts1}

\begin{promptbox}[title=Keyword Extraction Prompt]
Please take the following political claim and convert it into a set of a few (3-5) keywords that are optimized for searching Google.

The keywords will be used to look for relevant Reddit articles. Make sure that the keywords together carry the idea of what the claim is saying so that you can find relevant posts for the claim on Google. The keywords should retain the meaning of the claim, but in a more concise form.

Claim: ...

Only provide the keywords, without any other text. 
\end{promptbox}

\begin{promptbox}[title=Document Labeling Prompt]
You are an assistant that assesses the agreement between a DOCUMENT and a CLAIM.

CLAIM: ...

DOCUMENT: ...

Based on the DOCUMENT, do you agree with the CLAIM? Respond with your reasoning and then with either AGREE or DISAGREE.

If the DOCUMENT (even indirectly) supports the given CLAIM, respond with "AGREE". If the document (even indirectly) contradicts the CLAIM, respond with "DISAGREE".
If there is no support (indirect or direct) against or for the claim, output unrelated.
\end{promptbox}

\subsection{Evaluation}\label{sec:prompts2}

In all settings, two-shot examples—one true and one false claim from \textsc{RAGuard} training data—are provided in the context.

\begin{promptbox}[title=Zero-Context Prediction Prompt]
You are an expert fact-checker. Given a claim and some context, your task is to assess its accuracy and classify it as either TRUE or FALSE.

Examples:

Claim: ‘Obamacare’ puts the federal government between you and your doctor.

Output: FALSE

Claim: \$500 billion of the extension of the Bush tax cuts for the wealthy goes to 120,000 families.

Output: TRUE

Respond only with the following format, nothing else:

Output: [TRUE or FALSE]

Do not include any additional text, headers, or explanations outside this format.

Claim: ...
\end{promptbox}

\begin{promptbox}[title=Context-Based Prediction Prompt]
You are an expert fact-checker. Given a claim and some context, your task is to assess its accuracy and classify it as either TRUE or FALSE. The context is NOT ALWAYS relevant or correct. You must use your own judgement and must return a response.

Examples:

Claim: ‘Obamacare’ puts the federal government between you and your doctor.

Output: FALSE

Claim: \$500 billion of the extension of the Bush tax cuts for the wealthy goes to 120,000 families.

Output: TRUE

Respond only with the following format, nothing else:

Output: [TRUE or FALSE]

Do not include any additional text, headers, or explanations outside this format.

Context: ...

Claim: ...
\end{promptbox}

\section{Expanded Results and Analyses}\label{sec:analysis}

Table~\ref{tab:main2} is a version of the results table (Table\ref{tab:main}) with exact relative percent decreases shown.

\begin{table*}[h]
\centering
\resizebox{\textwidth}{!}{%
\begin{tabular}{lllllllll}
\toprule
& \multicolumn{3}{c}{Open Source} & \multicolumn{3}{c}{Closed Source} & \multicolumn{2}{c}{Reasoning} \\
\cmidrule(lr){2-3} \cmidrule(lr){4-6} \cmidrule(lr){7-8}
& OLMo-1B & Llama 3 & Mistral & Gemini 1.5 & GPT-4o & Claude 3.5 & DeepSeek & o4-mini \\
\midrule
Zero-Context Prediction & 
56.87 & 62.50 & 63.97 & 
61.06 & 67.33 & 74.51 & 
69.98 & 63.67 \\
RAG-1 & 
52.68$_{\textcolor{red}{\downarrow 7.4\%}}$ &
59.40$_{\textcolor{red}{\downarrow 5.0\%}}$ & 
59.14$_{\textcolor{red}{\downarrow 7.6\%}}$ & 
56.68$_{\textcolor{red}{\downarrow 7.2\%}}$ & 
64.80$_{\textcolor{red}{\downarrow 3.8\%}}$ & 
70.09$_{\textcolor{red}{\downarrow 5.9\%}}$ & 
66.88$_{\textcolor{red}{\downarrow 4.4\%}}$ & 
62.76$_{\textcolor{red}{\downarrow 1.4\%}}$ \\
RAG-5 & 
49.74$_{\textcolor{red}{\downarrow 12.5\%}}$ &
61.37$_{\textcolor{red}{\downarrow 1.8\%}}$ & 
58.91$_{\textcolor{red}{\downarrow 7.9\%}}$ & 
57.59$_{\textcolor{red}{\downarrow 5.7\%}}$ & 
65.90$_{\textcolor{red}{\downarrow 2.1\%}}$ & 
68.58$_{\textcolor{red}{\downarrow 8.0\%}}$ & 
57.81$_{\textcolor{red}{\downarrow 17.4\%}}$ & 
63.14$_{\textcolor{red}{\downarrow 0.8\%}}$ \\
Oracle Retrieval (All) & 
53.89$_{\textcolor{red}{\downarrow 5.2\%}}$ &
61.09$_{\textcolor{red}{\downarrow 2.3\%}}$ & 
51.55$_{\textcolor{red}{\downarrow 19.4\%}}$ & 
52.38$_{\textcolor{red}{\downarrow 14.2\%}}$ & 
53.22$_{\textcolor{red}{\downarrow 21.0\%}}$ & 
52.56$_{\textcolor{red}{\downarrow 29.5\%}}$ & 
50.06$_{\textcolor{red}{\downarrow 28.5\%}}$ & 
51.88$_{\textcolor{red}{\downarrow 18.5\%}}$ \\
Oracle Retrieval (Misleading) & 
44.04$_{\textcolor{red}{\downarrow 22.6\%}}$ &
36.81$_{\textcolor{red}{\downarrow 41.1\%}}$ & 
26.88$_{\textcolor{red}{\downarrow 58.0\%}}$ & 
30.57$_{\textcolor{red}{\downarrow 49.9\%}}$ & 
45.97$_{\textcolor{red}{\downarrow 31.7\%}}$ & 
35.98$_{\textcolor{red}{\downarrow 51.7\%}}$ & 
38.25$_{\textcolor{red}{\downarrow 45.3\%}}$ & 
33.39$_{\textcolor{red}{\downarrow 47.6\%}}$ \\
\bottomrule
\end{tabular}
}
\caption{Accuracy (\%) of various LLM backbones in RAG setup across three tasks and five evaluation settings. Subscripts indicate percent decrease from the Zero-Context baseline.}
\label{tab:main2}
\end{table*}

\subsection{GPT-4's Role in Annotation vs. Evaluation} \label{sec:gpt4role}
Although GPT-4 was used during data construction to guide labeling, it does not necessarily have an inherent advantage on our benchmark. When used to label documents, GPT-4 does not directly classify documents as misleading or not. Instead, it produces a “test” prediction for each claim (i.e., \textit{true} or \textit{false}) given the retrieved documents, which is then compared against the dataset gold verdicts to derive dataset labels (see Section~\ref{sec:llm-annotation}). Consequently, GPT-4 can still be misled by the same evidence when re-evaluated in the benchmark setting. GPT-4 does not necessarily know when evidence is misleading or not.

Because the dataset is partially constructed around GPT-4’s own failures, GPT-4 is theoretically expected to perform worse on these misleading instances because they exploit its prior weaknesses. However, GPT-4 does not drastically underperform compared to models like DeepSeek. This suggests that the misleading patterns in the dataset are not uniquely adversarial to GPT-4 but rather reflect broader challenges that also affect other models.

We hypothesize that the fact that GPT-4 does not drastically underperform other models is because of its inherent strength as a model. While parts of the dataset may be harder for GPT-4 due to its role in data construction, GPT-4 also remains a very strong model at fact-checking relative to other models tested. Recent fact-checking papers benchmarking LLMs find GPT-4 to improve the best of the LLMs tested \cite{tang2024minicheck}. This reasoning strength likely offsets the disadvantage, leading to competitive performance despite the dataset being partially shaped by its own failure cases.

\subsection{Retrieval Performance}\label{sec:retrievalappendix}
Retrieval performance is a standard metric in RAG benchmarks, but our dataset focuses on how models handle misleading or conflicting evidence. High retrieval accuracy alone does not ensure reliable answers due to misleading information in the corpus. Nonetheless, to provide a full view of system behavior, we report both conventional retrieval metrics and a tailored measurement called Misleading Retrieval Recall.

\begin{wrapfigure}{r}{0.6\linewidth}  
    \centering
    
    \includegraphics[width=0.5\textwidth]{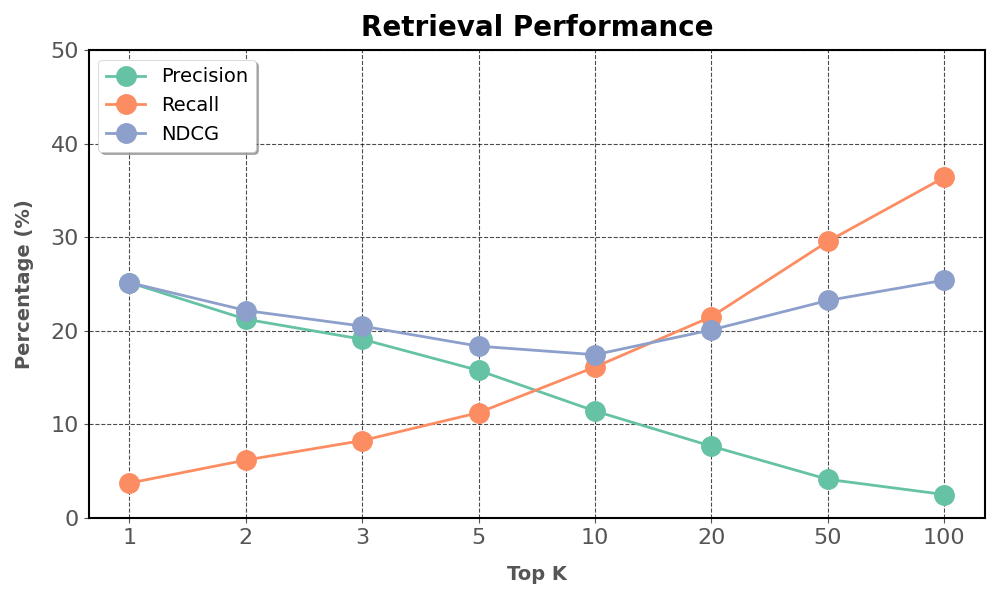}
       \caption{Retrieval Accuracy, Recall, and NDCG at Different Top K Levels}
    \label{fig:retrieval_metrics_line}
\end{wrapfigure}


Figure~\ref{fig:retrieval_metrics_line} shows Retrieval Precision, Recall, and Normalized Discounted Cumulative Gain (NDCG) for Task 2 (Standard RAG). Recall naturally rises with $K$, the number of documents that are returned as the highest-ranked results, while precision decreases. NDCG follows a non-monotonic trend, dipping around $K=10$ before recovering due to relevant items being unevenly distributed across ranked positions, causing reordering as $K$ changes.

We also report Misleading Retrieval Recall—the fraction of claims retrieving at least one misleading document. Zero-Context Baseline scores 0\%, while Oracle Retrieval scores 100\%. RAG-1 scores 21.3\%, increasing to 44.8\% for RAG-5, showing a higher risk of retrieving misleading content when retrieving more documents. As seen in Table~\ref{tab:main}, this correlates with lower overall accuracy.

\section{Qualitative Examples and Error Analysis}\label{sec:qualitative}
\subsection{Failure Analysis}\label{sec:failure}
Our analysis reveals two primary failure modes that explain how misleading evidence interferes with LLM reasoning. First, models often misinterpret subjective language or rhetorical tone as factual support. For example, a user’s frustrated question, “Is it normal to be taxed this much?” is incorrectly used to verify a claim about a tax hike. 

Second, models frequently latch onto superficial details such as numbers or names while overlooking broader context. In one case, a document referencing a “300 million dollar” cost unrelated to the target claim was used to support a “300 billion dollar” fiscal figure. These examples demonstrate that LLMs overweight surface-level signals and struggle with nuanced reasoning tasks like assessing temporal context or implicit framing, limitations that explain their vulnerability to misleading retrieval.

\begin{figure*}
    \centering
    \includegraphics[width=\textwidth]{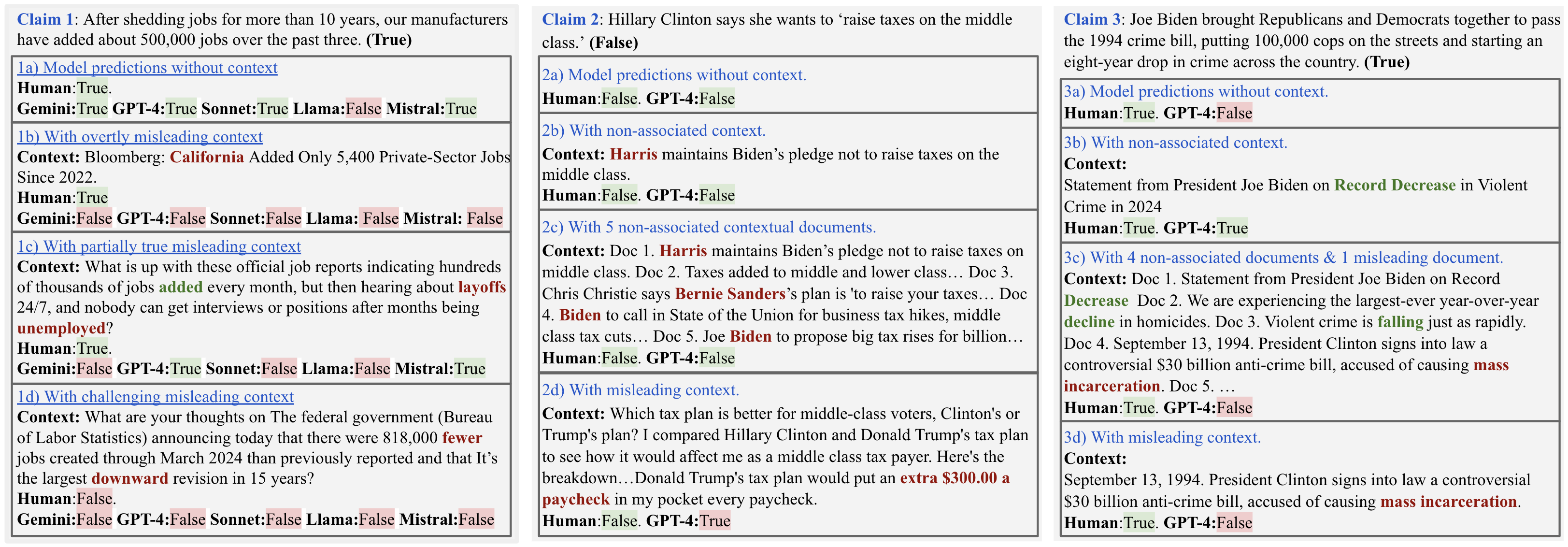}
    \caption{Example predictions on \textsc{RAGuard}, compared to the expected human response. Note that each column compares different prediction scenarios based on varying retrieved contexts for the same claim rather than a multi-turn process. \textit{Left:} Each system's classification of a true claim with three progressively misleading documents.
    \textit{Middle:} GPT-4o-based system's classification of a false claim with one noisy non-associated document, many noisy non-associated documents, and a misleading document.
    \textit{Right:} GPT-4o-based system’s classification of a true claim with a supporting non-associated document, one misleading document along with other supporting non-associated documents, and a misleading document.}
   
    \label{fig:discussion}
\end{figure*}

\subsection{Types of Misleading Documents}
Figure~\ref{fig:discussion} presents example system predictions on \textsc{RAGuard}, illustrating the impact of misleading documents. The left example highlights how misleading documents negatively affect the classification of a true claim. While misleading documents generally degrade system performance compared to zero-shot predictions, their specific influence varies based on their complexity. We distinguish three categories of misleading documents:
\begin{enumerate}
\item Overtly Misleading Document: This category includes documents that are evidently misleading to humans but still lead to incorrect predictions by all RAG systems. For example, in Figure~\ref{fig:discussion}, the document falsely comparing California's job growth to the national average misleads all systems (1b), despite their correct zero-shot predictions (1a). This suggests a form of selective bias, where the systems prioritize the provided information simply because it is included in the prompt, even though the instructions explicitly caution against assuming its correctness.

\item Partially True Misleading Document: These documents contain partial truths, making it necessary to apply reasoning to recognize their misleading nature. For example, as shown in Figure~\ref{fig:discussion}, one document criticizes unemployment but also states that ``official job reports are reporting jobs added'' (1c). While this statement supports the claim that 500,000 jobs were added, the document’s overall tone suggests rising unemployment. However, this suggestion is more of an opinion than a fact. Some LLMs, such as GPT-4o and Mistral, were able to reason through this contradiction and classify the claim correctly.

\item Challenging Misleading Document: These documents present significant challenges, even for human annotators. For example, a claim referencing job growth in the 2000s is incorrectly classified because the RAG system retrieves data from 2024, which accurately reports lower job creation (1d). The temporal misalignment in retrieved documents presents a fundamental challenge in this dataset and task.
\end{enumerate}



The middle example demonstrates GPT-4o’s ability to filter out noise from retrieved documents that are not associated with the claim but could be considered misleading documents in our dataset (e.g., documents using the same phrasing but referring to different individuals, such as ``Harris'' instead of ``Clinton'' in example 2b). Even when five unrelated documents are retrieved (2c), GPT-4o remains robust. However, when presented with a misleading document from the dataset (2d), GPT-4o fails, reinforcing the dataset’s effectiveness in challenging model performance beyond conventional RAG noise. This further explains the lower accuracy observed in the Oracle Retrieval setting in our baseline experiments.

The right example shows how GPT-4o tends to assign disproportionate weight to misleading documents, allowing them to override even non-associated supporting evidence. In the example, a non-associated document that contains supporting information (3b) enables GPT-4o to correct its initially incorrect zero-shot prediction (3a). However, when a misleading document is retrieved alongside other non-associated supporting documents (3c), the system incorrectly classifies the claim, similar to its behavior when only the misleading document is retrieved (3d). This demonstrates that misleading documents can have a stronger influence on the model’s classification, regardless of the presence of supporting evidence, highlighting a significant vulnerability in RAG systems.

These examples highlight three key findings. First, LLMs remain highly susceptible to misleading documents, even when their content is transparently incorrect. Second, misleading documents retrieved from the dataset exert a stronger influence than non-associated documents retrieved erroneously. Third, when misleading documents are present, they can significantly outweigh supporting evidence, leading to incorrect predictions. These findings emphasize the strength and uniqueness of our dataset in evaluating and challenging RAG-based model performance.

\section{Human Study Setup}\label{sec:humaninstruct}
\begin{figure}[h]
  \centering
  \fbox{\includegraphics[width=\textwidth]{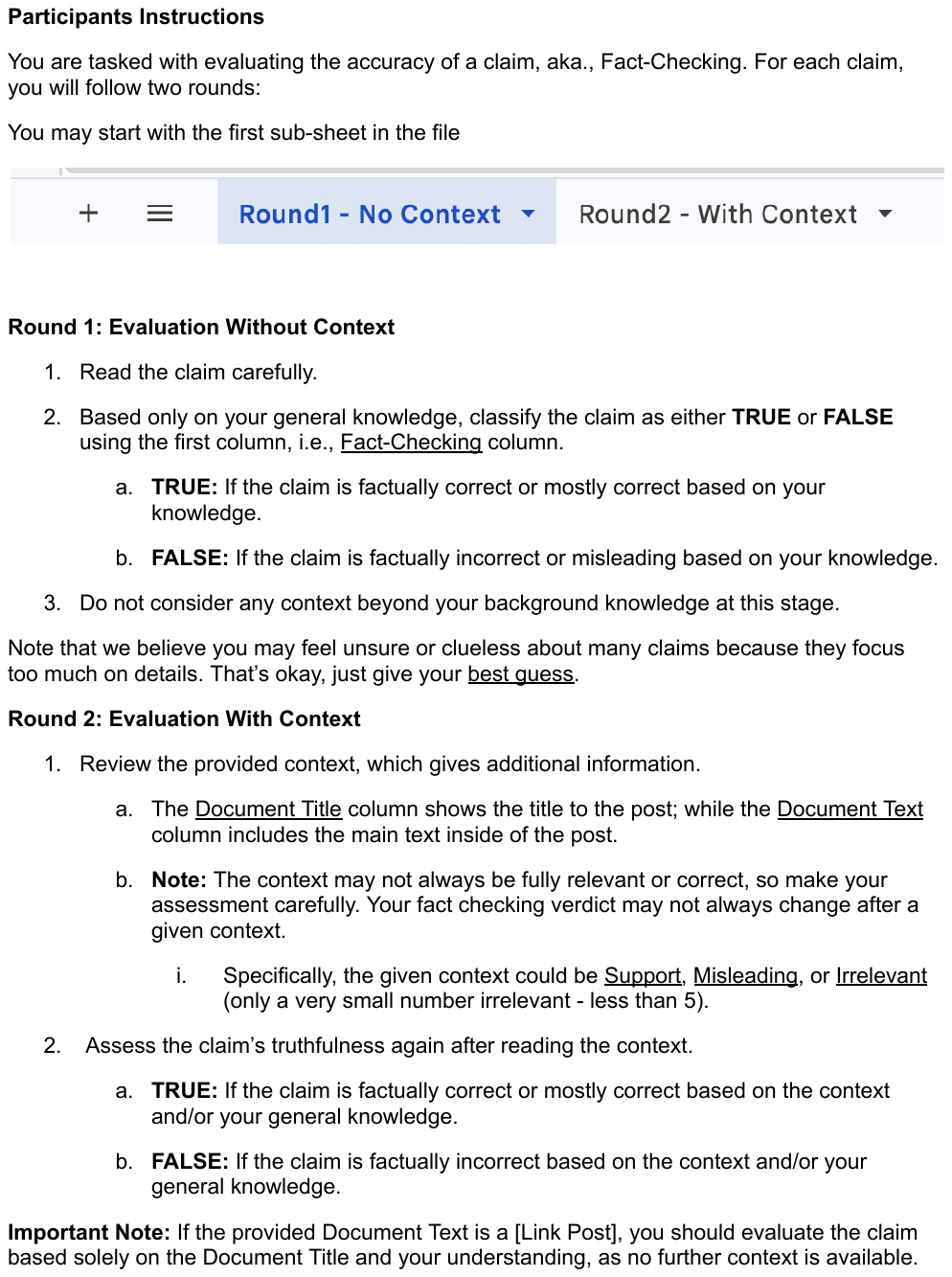}}
  \caption{Instructions provided to humans to verify a subset of the dataset.}\label{fig:annotators}
\end{figure}

We recruit four undergraduate and PhD students familiar with the U.S. political landscape. Annotators first make zero-context predictions, then are given one document without its label, and make predictions on the same claims, simulating the Oracle Retrieval setting. The annotators come from diverse backgrounds, both international and Californian, to reflect a range of political awareness. Figure~\ref{fig:annotators} displays instructions, which are provided along with a spreadsheet to complete the predictions.

We construct a 64-instance subset (balanced across true and false claims) representative of the main contributions over our full dataset, with 20 misleading, 36 supporting, and 8 unrelated documents. While the misleading and supporting document distribution reflecting the distribution of misleading and supporting documents in the full dataset, we down-sample unrelated documents, which are less impactful to model performance.

\end{document}